\documentclass[10pt,twocolumn,letterpaper]{article}

\usepackage{wacv}              % To produce the CAMERA-READY version
\newcommand{\method}[1]{\textsc{#1}}
\newcommand{\ours}{\method{Baller120}\xspace}
\usepackage{amssymb}
\usepackage{amsmath,amsfonts}
\usepackage{amsopn}
\usepackage{bm} 
\usepackage{graphicx}
\usepackage[font=small]{caption}
\usepackage{cuted}
\usepackage{booktabs} % For better looking tables
\usepackage{array}    % For column customization
\usepackage{makecell} % For multi-line cells
\usepackage{multirow}
\usepackage{adjustbox}
\usepackage{pifont}
\usepackage[table]{xcolor}
\usepackage[dvipsnames]{xcolor}
\usepackage{xspace}
\usepackage{tikz}
\usepackage{calc}
\usepackage{fontawesome5} 
\definecolor{oursrow}{HTML}{FFF7FF}
\definecolor{boxcolor}{HTML}{000bff}  
\definecolor{seqcolor}{HTML}{eb00b9}  

\newcommand{\flameicon}{\raisebox{-0.15em}{\includegraphics[height=1em]{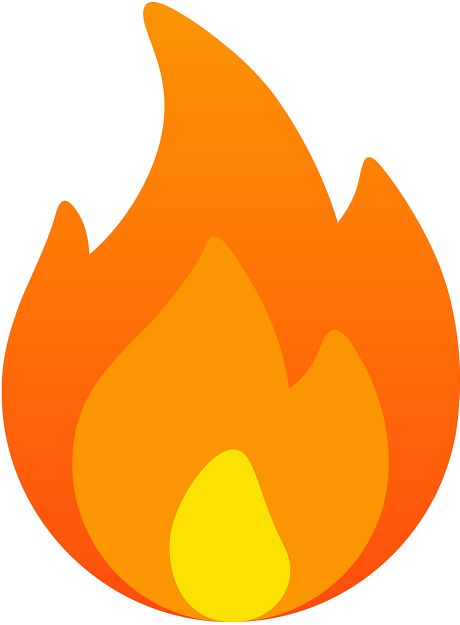}}}

\newcommand{\crossbox}{\raisebox{-0.15em}{\includegraphics[height=1em]{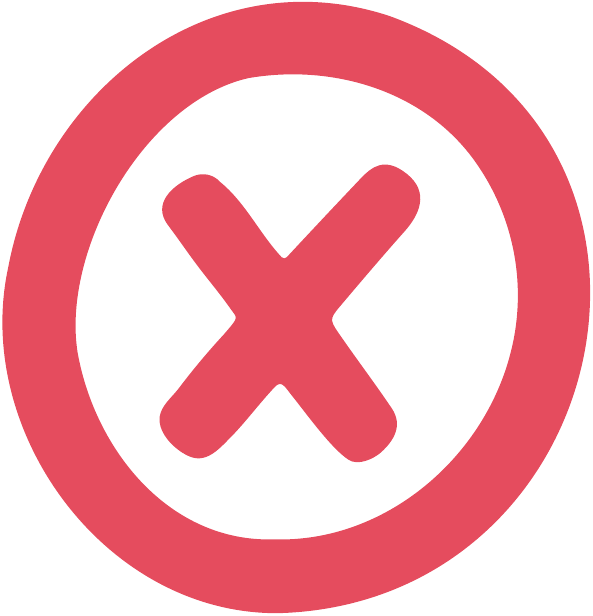}}}
\newcommand{\checkbox}{\raisebox{-0.15em}{\includegraphics[height=1em]{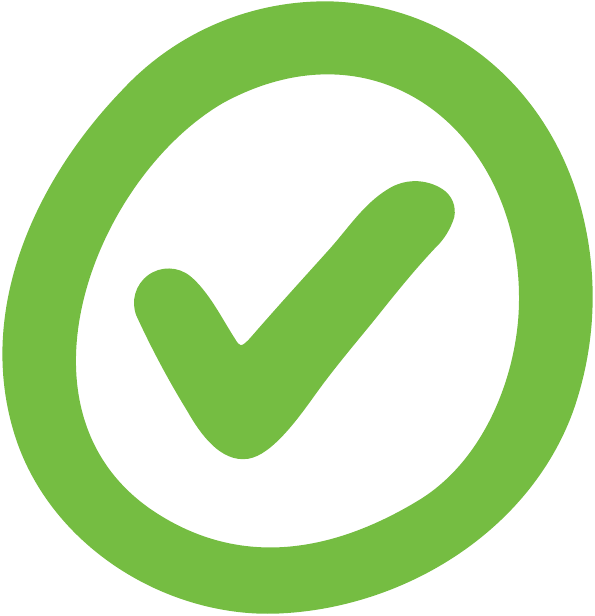}}}
\definecolor{rowsil}{HTML}{FFFEEC}  
\definecolor{rowskel}{HTML}{FFF2EC}  
\newcommand{\cs}{\cellcolor{rowsil}}
\newcommand{\ck}{\cellcolor{rowskel}} 
\definecolor{diffgreenbg}{RGB}{220,245,226}
\definecolor{diffgreenfg}{RGB}{20,130,60}
\definecolor{diffredbg}{RGB}{248,224,226}
\definecolor{diffredfg}{RGB}{150,55,60}
\definecolor{diffgraybg}{RGB}{235,235,235}
\definecolor{diffgrayfg}{RGB}{100,100,100}
\definecolor{rowblur}{HTML}{EFF9FF}
\newcommand{\cb}{\cellcolor{rowblur}}

\newcommand{\diffdown}[1]{%
  \makebox[\diffboxwidth][l]{%
  \tikz[baseline=(x.base)]{
    \node[
      fill=diffgreenbg,
      text=diffgreenfg,
      rounded corners=1.2pt,
      inner xsep=1.1pt,
      inner ysep=0.2pt
    ] (x) {\scriptsize\(\downarrow #1\)};
  }}%
}

\newcommand{\diffup}[1]{%
  \makebox[\diffboxwidth][l]{%
  \tikz[baseline=(x.base)]{
    \node[
      fill=diffredbg,
      text=diffredfg,
      rounded corners=1.2pt,
      inner xsep=1.1pt,
      inner ysep=0.2pt
    ] (x) {\scriptsize\(\uparrow #1\)};
  }}%
}

\newcommand{\diffsame}[1]{%
  \makebox[\diffboxwidth][l]{%
  \tikz[baseline=(x.base)]{
    \node[
      fill=diffgraybg,
      text=diffgrayfg,
      rounded corners=1.2pt,
      inner xsep=1.1pt,
      inner ysep=0.2pt
    ] (x) {\scriptsize\(-#1\)};
  }}%
}

\definecolor{wacvblue}{rgb}{0.21,0.49,0.74}
\definecolor{cvprpink}{RGB}{199,21,133}
\usepackage[
    pagebackref,
    breaklinks,
    colorlinks,
    linkcolor=red,
    citecolor=wacvblue,
    urlcolor=RubineRed
]{hyperref}

\def\wacvPaperID{667} % *** Enter the WACV Paper ID here
\def\confName{WACV}
\def\confYear{2027}

\title{Probing Identity-Specific Motion Signatures: A Controlled Diagnostic Study}
\author{
Yingtie Lei\thanks{Equal contributions.}\;$^1$ \quad
Fangxun Liu\footnotemark[1]\;$^1$ \quad
Baicheng Wu\footnotemark[1]\;$^1$ \quad
Colin Lee$^1$ \quad
Ziheng Zhang$^1$ \quad
Junke Yang$^1$  \\
Zhiyuan Tao$^1$ \quad
Xuyan Huang$^1$ \quad
Shuheng Wang$^1$ \quad
William Koran$^1$ \quad
Kyle Park$^1$ \\
Elijah H. Buckwalter$^1$ \quad
Cheng-Hsuan Chiang$^1$ \quad
Tejas Naik$^1$ \quad
Daniel Yi$^1$ \quad
Wei-Lun Chao\thanks{Corresponding author.}\;$^{1,2}$ \\[1em]
$^1$The Ohio State University \quad
$^2$Boston University
}

\begin{document}
\maketitle

%%%%%%%%% MAIN BODY
\begin{strip}
\vskip -60pt
\centering
\includegraphics[width=0.95\linewidth]{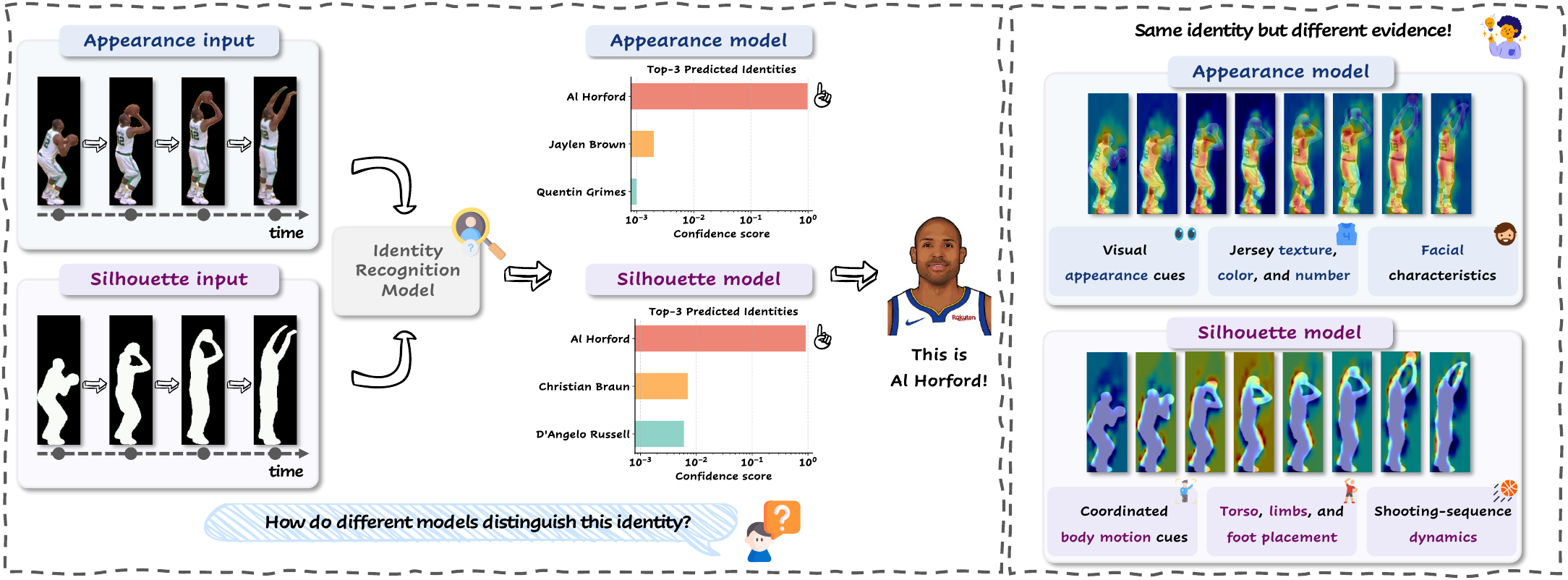}
\vskip -5pt
\captionof{figure}{Overview of our diagnostic probe for identity-specific motion signatures. Given the same free-throw sequence, models can rely on appearance cues, such as jersey texture, color, number, and facial characteristics, or on appearance-suppressed silhouettes that preserve motion dynamics. By contrasting appearance and silhouette models, we examine \textit{\textbf{whether modern video models move beyond static visual shortcuts and capture human-interpretable motion patterns that serve as identity-specific signatures}}.}
\label{fig:teaser}
%\vskip -10pt
\end{strip}
\begin{abstract}
Identity recognition (e.g., person, animal re-identification) has traditionally relied heavily on static appearance cues. Yet motion---consistent, individual-specific dynamics---can provide a complementary and potentially more robust signature, especially when appearance is weak or variable. 
%For example, even without a clear view of a player's face or jersey, observers can often recognize athletes from characteristic pitching, shooting, or running styles. 
This raises a fundamental question: \textbf{when identity-specific motion cues are clearly present, to what extent do modern video models use them for recognition?} To investigate this question, we conduct a systematic diagnostic study and introduce \textbf{\ours}, a controlled benchmark of 120 professional basketball players performing free-throws. By focusing on the same multi-phase action across individuals, \ours reduces action-level variation and identity-correlated acquisition biases, enabling fine-grained analysis of identity-specific kinematic patterns. We find that modern video models can predict identity accurately from RGB videos, but often rely on static appearance cues such as faces and jersey regions, even when informative motion cues are available. Strikingly, when appearance is suppressed through silhouette-only or skeleton-only inputs, the same model architectures shift toward motion micro-patterns (e.g., foot placement and elbow bending). Despite containing less visual information, appearance-suppressed representations achieve competitive accuracy and stronger robustness to appearance shifts. Our qualitative analyses further show that appearance-suppressed models attend to distinctive motion patterns across individuals. Overall, our study demonstrates that identity-specific motion signatures are present, informative, and learnable, but modern video models may overlook them in favor of easier static shortcuts unless appearance cues are explicitly suppressed.
\end{abstract}

\section{Introduction}
\label{sec:intro}
Visual recognition has advanced rapidly, progressing from coarse category recognition to fine-grained subtype classification and individual identity recognition~\cite{deng2009imagenet, krizhevsky2012imagenet, ye2021deep, wei2021fine}. Modern deep learning models now achieve strong performance across a wide range of recognition tasks~\cite{radford2021learning, liu2022convnet, zhai2023sigmoid, wang2025internvideo2, simeoni2025dinov3}. Beyond enabling practical applications, this progress raises a deeper scientific opportunity: \emph{accurate models may reveal visual or behavioral traits that are difficult for humans to explicitly characterize.} This possibility is especially intriguing for identity recognition, where the cues that distinguish one individual from another are often subtle and fine-grained. 

One such cue is motion. Humans and animals exhibit consistent individual-specific dynamics that can persist across contexts~\cite{filipi2022gait, shen2024comprehensive}. Unlike static appearance, which may be ambiguous, degraded, or altered over time, motion can reflect biomechanical habits that are harder to change or disguise. For example, observers can often recognize athletes from their pitching, shooting, or running styles even without a clear view of the face or jersey. This makes video-based identity recognition promising for uncovering subtle and previously unknown motion signatures.

However, before this promise can be realized, we must first ask a more basic diagnostic question: \emph{do video models actually use motion patterns for identity recognition?} High recognition accuracy alone does not show that a model uses motion. It may instead rely on static cues, such as faces, clothing, or body patterns, or on incidental correlations introduced during data collection. In unconstrained video collections, identity may be correlated with nuisance factors such as action type, skill level, capture location, camera viewpoint, or environmental context. A model may thus recognize identities by exploiting dataset-specific shortcuts rather than learning how an individual moves.

This motivates a \emph{controlled} setting in which identity-specific motion cues are clearly present and interpretable, while identity-correlated acquisition biases are reduced. We curate \textbf{\ours}, a controlled diagnostic dataset designed to probe motion signatures in identity recognition. \ours contains over $4{,}500$ short \emph{free-throw} clips of 120 professional basketball players, collected across diverse games and broadcast camera viewpoints. Free-throws provide a useful testbed because they are structured, multi-phase actions shared across individuals: each sequence includes preparation, upward motion, ball release, and follow-through. Focusing on professional players further reduces coarse variation due to skill level, allowing us to study subtle but recognizable identity-specific differences in posture, timing, and body coordination. To further separate motion from static appearance, we provide frame-wise silhouette masks and body skeletons for each sequence, suppressing visual cues such as facial details, jersey numbers, and logos and encouraging models to rely more on \emph{how players move}.

%{\color{orange}A little bit concerned about the word ``must'', maybe ``encouraging models to rely more on appearance-suppressed kinematic and configuration cues.''} 

Using \ours, we evaluate whether modern video backbones actually rely on motion when repurposed for identity recognition. We study MViTv2~\cite{li2022mvitv2}, VideoMAEV2~\cite{wang2023videomae}, and UniFormerV2~\cite{li2023uniformerv2}, originally developed for action recognition, under three input regimes: full-appearance, silhouette-only, and skeleton-only videos. For each regime, we fine-tune a separate model and evaluate it on held-out clips, allowing us to compare how recognition changes as static appearance cues are reduced.

As expected, models trained on full-appearance videos recognize players accurately, showing the strong transferability of modern pre-trained video backbones. More importantly, silhouette-only and skeleton-only models also achieve competitive accuracy, indicating that appearance-suppressed videos retain sufficient information for identity recognition. To examine what cues drive these predictions, we evaluate models under a split where each player’s static appearance distribution differs between training and testing. Full-appearance models degrade sharply under this shift, while appearance-suppressed models are more robust. Since full-appearance videos contain the same motion information available to silhouettes and skeletons, this contrast suggests that they do not lack motion cues; rather, they preferentially rely on static cues that fail to transfer across appearance changes. CAM-based saliency analysis~\cite{zhou2016learning} further supports this finding. When appearance is suppressed, models attend to motion micro-signatures that shift across body parts as the free-throw routine unfolds. These cues are stable across games and camera viewpoints for the same player, yet differ systematically across players performing the same action. In contrast, full-appearance models, despite having access to motion, concentrate largely on static regions such as faces and jerseys. These results suggest that motion cues are informative and identity-linked, but can be overshadowed by easier appearance shortcuts.

We further ask whether explicit appearance suppression is always necessary for learning motion signatures. To test whether this behavior is specific to identity recognition, we replace identity prediction with action discrimination, distinguishing three-pointers from free-throws using the same video backbones. In this setting, even full-appearance models attend to motion-relevant body regions rather than static appearance cues. This suggests a broader tendency: when optimized for accuracy, video models exploit the easiest predictive signal. For identity recognition, that signal is often static appearance; for action recognition, it is motion. Overall, our findings show that identity-specific motion signatures are present, learnable, and robust, but may be overlooked unless appearance cues are explicitly suppressed, pointing toward the need for video models that more reliably capture fine-grained individual dynamics.

% \noindent\textbf{Positioning.} Existing identity recognition benchmarks, including re-identification and gait recognition~\cite{zheng2017person,schneider2019past,zhu2021gait,zheng2022gait,li2023depth}, often emphasize recognition accuracy in unconstrained environments. Our objective is complementary: rather than pursuing generalization to unseen identities, actions, or domains, we ask what cues video models use when identity-specific motion cues are clearly present. We introduce a controlled benchmark and diagnostic protocol that reduce identity-correlated acquisition biases and separate static cues from motion-based identity cues. This controlled setting is a scientific design choice, not merely a limitation.

% free-throws provide a suitable testbed: players perform the same structured, multi-phase action at a comparable professional standard; each identity has enough clips to reveal within-identity regularities; and the motion patterns are interpretable. While gait recognition has established motion as an important identity cue, walking is only one form of individual motion, and unconstrained datasets may inadvertently tie identity to viewpoint, clothing, scene, or capture conditions. \ours complements prior benchmarks by focusing on skilled action, reducing incidental correlations, and exposing identity-specific motion signatures across interpretable phases. To our knowledge, \ours is the first controlled diagnostic dataset designed to make identity-specific motion signatures in skilled action discoverable and interpretable.
\begin{figure}[t]
    \centering
    \includegraphics[width=\linewidth]{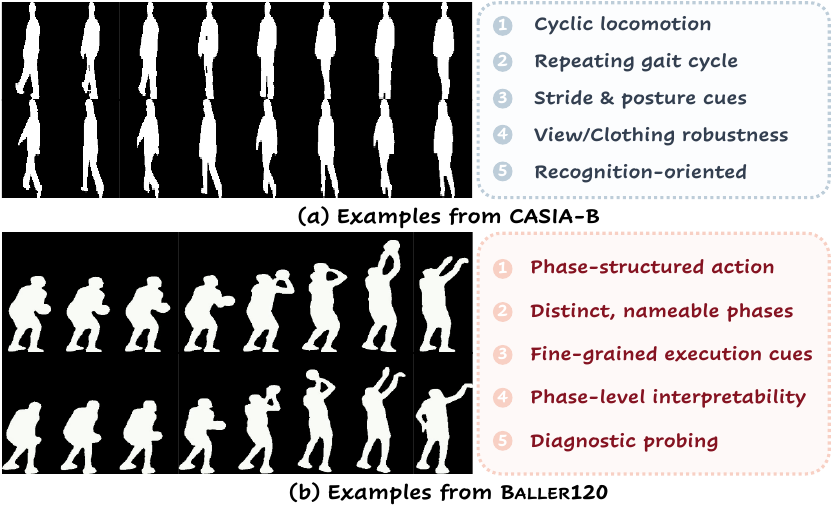}
    \vspace{-15pt}
    \caption{\textbf{Gait benchmarks vs.~\ours.} Each row shows a different individual. \ours is designed as a diagnostic probe: it focuses on free-throws, a multi-phase skilled action that naturally aligns execution across individuals, making it easier to visualize and verify the fine-grained identity-specific motion cues used by a model. In contrast, CASIA-B~\cite{yu2006framework} focuses on identity recognition from cyclic walking motion, where inter-identity differences can be harder to inspect directly. \ours therefore complements gait benchmarks as a diagnostic setting rather than a harder recognition benchmark.}
    %\caption{\textbf{Gait benchmarks vs.~\ours.} Each row contrasts a property of the two settings: gait (\eg, CASIA-B) captures a single cyclic motion curated for recognition, whereas \ours captures a phase-structured free-throw supporting fine-grained, human-verifiable diagnosis. \ours complements gait as a diagnostic probe, not a harder benchmark.}
    \label{fig:gait_vs_baller}
    \vspace{-16pt}
\end{figure}

\noindent\textbf{Positioning.} Existing identity recognition benchmarks, including re-identification and gait recognition~\cite{zheng2017person,schneider2019past,zhu2021gait,zheng2022gait,li2023depth}, often emphasize recognition accuracy in unconstrained environments. Our objective is complementary: rather than pursuing generalization to unseen identities, actions, or domains, we ask what cues video models use when identity-specific motion cues are clearly present. We introduce a controlled benchmark and diagnostic protocol that reduce identity-correlated acquisition biases and separate static cues from motion-based identity cues. This controlled setting is a scientific design choice rather than a limitation.
% ---it is what makes the diagnosis interpretable.

Free-throws provide a suitable testbed: players perform the same structured, multi-phase action at a comparable professional standard; each identity has enough clips to reveal within-identity regularities; and the motion patterns can be directly inspected. While gait recognition has established motion as an important identity cue, walking is only one form of cyclic motion. Moreover, unconstrained datasets may inadvertently tie identity to viewpoint, clothing, scene, or capture conditions. \ours complements prior benchmarks by focusing on skilled action, reducing incidental correlations, and exposing identity-specific motion signatures across interpretable phases (\ie, set, rise, release, follow-through); see \cref{fig:gait_vs_baller} for a comparison. To our knowledge, \ours is the first controlled diagnostic dataset designed to make identity-specific motion signatures in skilled action discoverable and verifiable.

\section{Related Work}
\label{sec:related}
\noindent\textbf{Datasets for Identity Recognition.} Existing datasets span three major settings. \textbf{(1) Person re-identification (Re-ID)} aims to match individuals across distinct observations. Image-based datasets, such as CUHK03~\cite{li2014deepreid}, Market-1501~\cite{zheng2015scalable}, and MSMT17~\cite{wei2018person}, match cropped pedestrians across camera views. Video-based datasets, including iLIDS-VID~\cite{wang2014person} and MARS~\cite{zheng2016mars}, use tracklets as the matching unit. To study stronger appearance variation, PRCC~\cite{yang2019person}, LTCC~\cite{qian2020long}, and DeepChange~\cite{xu2023deepchange} focus on clothing changes and long-term appearance shifts in image-based Re-ID, while CCVID~\cite{gu2022clothes} and MEVID~\cite{Davila_2023_WACV} extend these settings to videos. \textbf{(2) Animal Re-ID} goes beyond human subjects and supports wildlife monitoring by recognizing individuals from visual biometric patterns, such as stripes, spots, fins, or facial-scale markings. WildlifeReID-10k~\cite{adam2025wildlifereid}, for example, provides a large-scale multi-species benchmark with over 10,000 individuals across diverse capture conditions. \textbf{(3) Gait recognition} studies identity from walking patterns, typically using short video clips, with representative datasets including CASIA-B~\cite{yu2006framework}, GREW~\cite{zhu2021gait}, Gait3D~\cite{zheng2022gait}, and CCPG~\cite{li2023depth}. We provide a detailed comparison between these datasets and \ours in~\Cref{tab:dataset_comparison}.

\noindent\textbf{Biometrics for Identity Recognition.}
Physiological biometrics provide stable identity cues~\cite{calder2005understanding,daugman2009iris,pankanti2002individuality,han2003personal}, but often raise privacy concerns. Behavioral biometrics instead recognize individuals from how they move or act. Gait recognition shows walking patterns can carry identity-discriminative information across changes in viewpoint, clothing, and sensing condition~\cite{yu2006framework,zhu2021gait,zheng2022gait,li2023depth,wang2023dygait,ye2024biggait}. However, gait focuses on cyclic locomotion, a relatively regular form of motion. Richer actions contain more varied spatio-temporal structure, yet large-scale datasets such as Kinetics \cite{kay2017kinetics,carreira2018short,carreira2019short}, Moments-in-Time~\cite{monfort2019moments}, and Something-Something \cite{goyal2017something} focus on \textit{what action is performed}, not \textit{who performs it}. Whether fine-grained action dynamics can serve as identity-specific biometric cues remains underexplored.

\section{Diagnostic Protocol Design}
\label{sec:protocol}

%\noindent\textbf{Overview.}~We design our study as a \textit{diagnostic probe} rather than a general-purpose benchmark. As motivated in~\cref{sec:intro}, high accuracy alone does not show that a model relies on motion: identity can correlate with appearance, action-level variation, skill level, or data acquisition conditions, and a model may exploit any of these. The core challenge is that some cues are \textit{execution-independent}---a player's face or jersey identifies them no matter how they move---while others are \textit{execution-dependent}, reflecting how the action is carried out. 

\noindent\textbf{Overview.}~We study the question: \emph{when identity-specific motion cues are clearly present, to what extent do modern video models use them for identity recognition?} As discussed in~\cref{sec:intro}, high accuracy alone does not show that a model relies on motion: identity may correlate with appearance, action type, skill level, or data acquisition conditions. We therefore design \ours as a \textit{diagnostic probe} rather than a general-purpose benchmark. Our goal is to reduce dataset-specific biases and separate cues into two groups: \textit{execution-independent} cues, such as a player's face or jersey, which identify the player regardless of movement; and \textit{execution-dependent} cues, which reflect how the action is performed. This design allows us to suppress the former and inspect the latter. \ours implements this idea through four criteria: (\textit{i})~controlling coarse action-level variation, (\textit{ii})~suppressing execution-independent appearance and anatomy, (\textit{iii})~exposing execution-dependent evidence through complementary input regimes, and (\textit{iv})~supporting human-interpretable, phase-level attribution.

\noindent\textbf{Controlling Action and Context.} We focus on a single standardized action: the basketball free-throw, performed by all identities at a comparable professional level. Holding the action constant reduces coarse action- and scene-level differences common in unconstrained videos, while restricting to professional players narrows skill-level variation. This makes inter-person differences concentrate more on how each player executes the same routine. The shared multiple phases (set, rise, release, and follow-through) also make the setting diagnosable, allowing evidence to be compared at corresponding moments.

\noindent\textbf{Suppressing Execution-independent Appearance and Anatomy.} We reduce two types of cues that can identify a player regardless of motion. Static appearance, such as face, jersey, and texture, is removed through silhouette videos derived from frame-wise masks. Fixed anatomy, such as absolute scale and height, is reduced by spatially normalizing each sequence to a common scale; see~\cref{sec:dataset}. The remaining signal thus depends more on how the body moves than on static appearance or body size. We also consider skeletons from off-the-shelf pose estimators, which further abstract away appearance and body shape.

\noindent\textbf{Exposing Execution-dependent Evidence.} We examine the cues that remain after suppression: how each player executes the same routine. Some cues appear within a single phase, such as the set posture, release pose, or follow-through configuration. These are not fixed anatomy, since they depend on how the player performs the action. Other cues unfold over time, such as transitions between phases. Since no single representation captures all of these signals, we study complementary regimes: silhouettes preserve the dense body envelope, while skeletons preserve sparse joint positions and motion.
 %We therefore interpret identity cues through evidence shared across regimes, rather than through any single input alone.

\noindent\textbf{Supporting Human-interpretable Attribution.} We use the structure of the free-throw as a natural vocabulary for inspection. When attribution highlights a body region, it can be linked to an observable execution pattern within a specific phase. We therefore treat attribution as diagnostic support rather than standalone proof, and ask whether a model's evidence is stable within identity, distinct across identities, and aligned with interpretable phases.

\section{The \ours Dataset}
\label{sec:dataset}
\begin{figure}[t]
    \centering
    \includegraphics[width=0.9\linewidth]{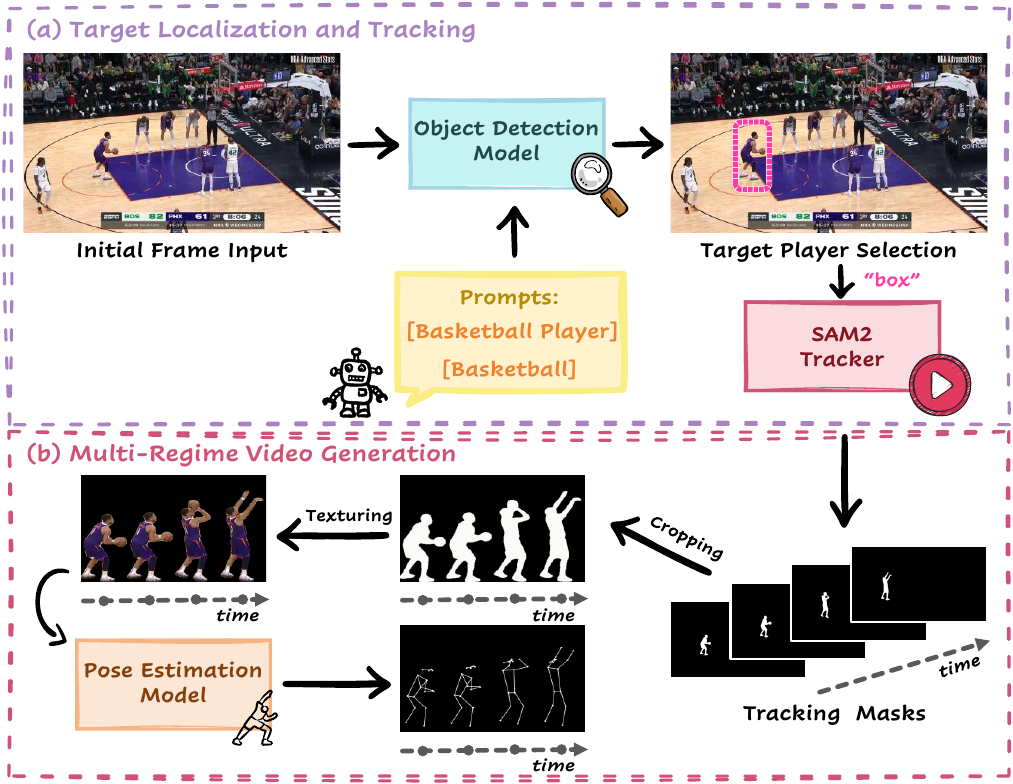}
    \vspace{-5pt}
    \caption{Data pipeline. (a) We localize all players in the first frame using Grounding DINO~\cite{liu2024grounding} prompted by text queries, manually select the target, and pass the box to SAM2~\cite{ravi2024sam2} to propagate frame-wise masks. (b) From the tracked masks, we generate three input regimes: appearance, silhouette, and skeleton.}
    \label{fig:dataset_construction}
    \vspace{-15pt}
\end{figure}
\noindent\textbf{Dataset Construction.} Following the diagnostic protocol in~\cref{sec:protocol}, we curate~\ours from official NBA broadcast replays, focusing on free-throw sequences. 
%As a ritualized, multi-phase routine shared by all players at a comparable standard, the free-throw ensures every identity undergoes the same phases, enabling cross-identity comparison while preserving subtle, identity-discriminative differences in execution. 
The pipeline is shown in~\cref{fig:dataset_construction}. Each clip is manually trimmed from the end of dribbling to shot release, and verified to ensure the target player remains visible and trackable throughout. Free-throw scenarios also naturally reduce player occlusion, yielding more reliable segmentation masks.

\noindent\textbf{Localizing and Tracking Target Players.} Given a trimmed clip, we feed the first frame into Grounding DINO~\cite{liu2024grounding}, prompted with text queries to localize all visible players. The target player is manually identified, and their bounding box is passed as a spatial prompt to SAM2~\cite{ravi2024sam2}, which propagates frame-wise segmentation masks across the entire sequence. To ensure temporal consistency, the basketball is included in the player mask prior to shot release and excluded afterward, so that the ball's trajectory is not encoded as part of the player's motion.

\noindent\textbf{Generating Multi-Regime Videos.} Using the frame-wise masks, we crop each clip around the player with a single union bounding box over all frames, and resize the crop to a common scale while preserving aspect ratio.
This removes per-frame jitter and absolute size while keeping the action's spatial range intact, reducing reliance on camera position and body scale, which are execution-independent. Within the cropped region, we derive three input regimes. \textit{Appearance videos} preserve the visual content inside the mask while removing background context.~\textit{Silhouette videos} retain only the binary mask, suppressing face, jersey, and texture while keeping how the body is configured and moves through the routine. \textit{Skeleton videos} render sparse body keypoints estimated from the appearance frames using RTMPose~\cite{jiang2023rtmpose}, discarding the dense body envelope while retaining joint configuration and temporal progression.

\begin{figure}[t]
    \centering
    \includegraphics[width=0.85\linewidth]{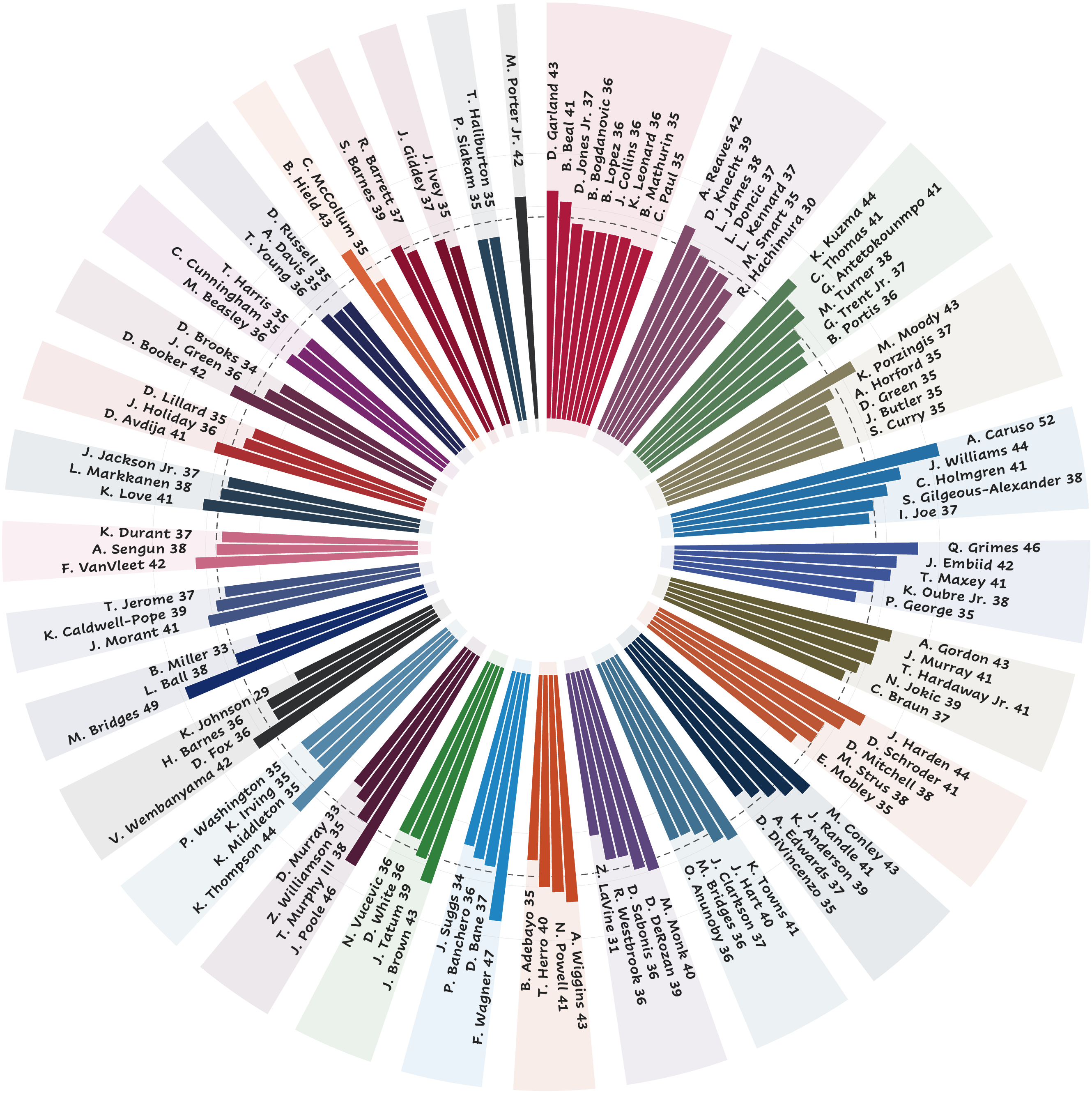}
    \vspace{-8pt}
    \caption{Per-identity clip distribution across the 120 NBA players in~\ours. Colors denote NBA teams; numbers and bar lengths indicate clip counts per player.}
    \label{fig:dataset_distribution}
    \vspace{-10pt}
\end{figure}

\noindent\textbf{Dataset Statistics.} As illustrated in~\cref{fig:dataset_distribution},~\ours comprises 4,583 free-throw sequences across 120 professional NBA players from all 30 teams, averaging roughly 38 clips per identity, collected across games, viewpoints, and jersey appearances. Each clip captures a complete free-throw averaging 1.71 seconds ($\sim$2.18 hours total), and is provided in appearance, silhouette, and skeleton regimes.

% {\color{blue}Although \ours is smaller than large-scale application-driven Re-ID benchmarks, its scale is appropriate for a controlled diagnostic study: it includes 120 identities with roughly 40 clips per identity, collected across games, viewpoints, and jersey appearances. This design prioritizes repeated observations under a shared action protocol, which is critical for exposing identity-specific regularities and isolating identity-relevant motion signatures.}
\begin{table}[t]
    \centering
    \caption{\textbf{Comparison of \ours with representative identity-recognition datasets.} \ours complements Re-ID and gait benchmarks by focusing on a multi-phase skilled action with paired appearance and appearance-suppressed video regimes, enabling diagnostic analysis of identity-relevant execution cues.}
    \vspace{-7pt}
    \label{tab:dataset_comparison}
    \setlength{\tabcolsep}{4pt}
    \begin{adjustbox}{width=\linewidth}
    \begin{tabular}{l c c c c c c}
        \toprule
        \multirow{2}{*}{\textbf{Dataset}} 
        & \multicolumn{2}{c}{\textbf{Scale}} 
        & \textbf{Data} 
        & \textbf{Action} 
        & \textbf{App.} 
        & \textbf{Explain.} \\
        \cmidrule(lr){2-3}
        & \textbf{\#IDs} 
        & \textbf{\textcolor{boxcolor}{\#Box}/\textcolor{seqcolor}{\#Seq}} 
        & \textbf{Type} 
        & \textbf{Protocol} 
        & \textbf{Suppression} 
        & \textbf{Support} \\
        \midrule
        \multicolumn{7}{c}{\textit{\textbf{Person Re-identification}}} \\
        \midrule
        CUHK03~\cite{li2014deepreid}         & $1{,}360$  & \textcolor{boxcolor}{$13{,}164$}   & Image & None & \crossbox & \crossbox \\
        Market-1501~\cite{zheng2015scalable} & $1{,}501$  & \textcolor{boxcolor}{$32{,}668$}   & Image & None & \crossbox & \crossbox \\
        MSMT17~\cite{wei2018person}          & $4{,}101$  & \textcolor{boxcolor}{$126{,}441$}  & Image & None & \crossbox & \crossbox \\
        PRCC~\cite{yang2019person}           & $221$      & \textcolor{boxcolor}{$33{,}698$}   & Image & None & \crossbox & \crossbox \\
        LTCC~\cite{qian2020long}             & $152$      & \textcolor{boxcolor}{$17{,}138$}   & Image & None & \crossbox & \crossbox \\
        DeepChange~\cite{xu2023deepchange}   & $1{,}121$  & \textcolor{boxcolor}{$178{,}407$}  & Image & None & \crossbox & \crossbox \\
        \midrule
        iLIDS-VID~\cite{wang2014person}      & $300$      & \textcolor{seqcolor}{$600$}        & Video & None & \crossbox & \crossbox \\
        MARS~\cite{zheng2016mars}            & $1{,}261$  & \textcolor{seqcolor}{$20{,}715$}   & Video & None & \crossbox & \crossbox \\
        CCVID~\cite{gu2022clothes}           & $226$      & \textcolor{seqcolor}{$2{,}856$}    & Video & None & \crossbox & \crossbox \\
        MEVID~\cite{Davila_2023_WACV}        & $158$      & \textcolor{seqcolor}{$8{,}092$}    & Video & None & \crossbox & \crossbox \\
        \midrule
        \multicolumn{7}{c}{\textit{\textbf{Gait Recognition}}} \\
        \midrule
        CASIA-B~\cite{yu2006framework}       & $124$      & \textcolor{seqcolor}{$13{,}640$}   & Video & Cyclic gait & \checkbox & \crossbox \\
        GREW~\cite{zhu2021gait}              & $26{,}345$ & \textcolor{seqcolor}{$128{,}671$}  & Video & Cyclic gait & \checkbox & \crossbox \\
        Gait3D~\cite{zheng2022gait}          & $4{,}000$  & \textcolor{seqcolor}{$25{,}309$}   & Video & Cyclic gait & \checkbox & \crossbox \\
        CCPG~\cite{li2023depth}              & $200$      & \textcolor{seqcolor}{$16{,}566$}   & Video & Cyclic gait & \checkbox & \crossbox \\
        \midrule
        \rowcolor{oursrow}
        \textbf{\ours}                       & $120$      & \textcolor{seqcolor}{$4{,}583$}    & Video & Multi-phase skilled & \checkbox & \checkbox \\
        \bottomrule
    \end{tabular}
    \end{adjustbox}
    \vspace{-12pt}
\end{table}

% \noindent\bluebold{Dataset size.~(@\Rtwo)}~Our goal is a controlled diagnostic setting rather than a large-scale benchmark.~For a \textbf{Concept \& Feasibility} contribution, we humbly believe the scale is appropriate: 120 identities are comparable to CCPG's 200, with roughly 40 clips per player across games, viewpoints, and jersey types. The constrained free-throw setting helps reduce confounders and isolate identity-relevant motion signatures. We agree that larger-scale validation is important future work and will clarify this limitation.

\noindent\textbf{Dataset Comparison.} \Cref{tab:dataset_comparison} situates \ours among representative Re-ID and gait datasets, which are typically curated for recognition accuracy in unconstrained or view-varying settings. In contrast, \ours is built for diagnosis: it asks which cues a model uses, not only how accurately it recognizes. This goal is reflected in the action design. Re-ID datasets leave actions unconstrained, and gait datasets focus on cyclic locomotion, whereas \ours fixes a multi-phase skilled action shared across identities, making inter-person variation more attributable to individual execution than to action type or scene. \ours also pairs each sequence with appearance-suppressed silhouette and skeleton regimes, whose cues can be inspected phase by phase to support interpretable diagnosis. These choices explain its scale: with 120 identities, roughly 38 clips per identity, and dense annotations, \ours is smaller than large-scale Re-ID benchmarks but comparable to gait datasets such as CASIA-B and CCPG. \ours therefore complements existing benchmarks by filling a gap they were not designed to address.

\section{Repurposing Action-Recognition Backbones for Identity Probing}
\begin{figure}[t]
\centering
% \vspace{-18pt}
\includegraphics[width=\linewidth]{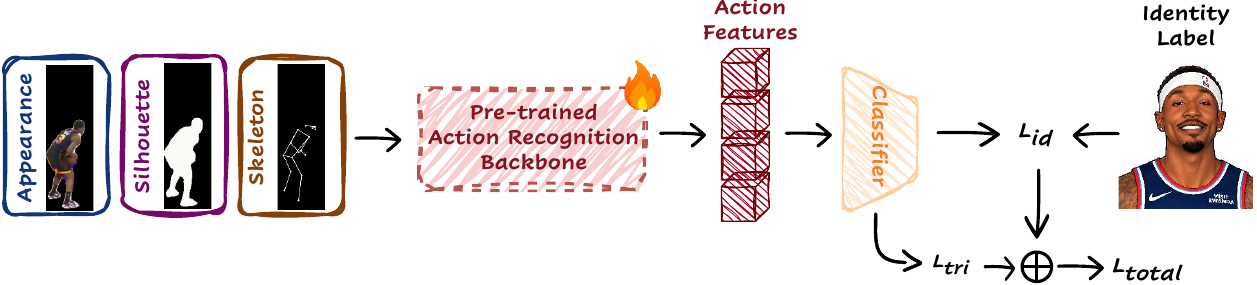}
\caption{\textbf{Identity probing with action-recognition backbones.} Each video, in one of three input regimes (appearance, silhouette, or skeleton), is passed through a fully fine-tuned (\flameicon) backbone, whose embedding is optimized with an identity classification loss $\mathcal{L}_{id}$ and a triplet loss $\mathcal{L}_{tri}$.}
\vspace{-15pt}
\label{fig:ar-to-identity}
\end{figure}
\label{sec:ar_to_identity}
Action recognition (AR) and identity recognition pursue different objectives:~AR classifies \textit{what action occurs}, whereas identity recognition determines \textit{who an individual is}, often from static appearance. Since all our clips depict the same free-throw, coarse action-category differences cannot explain identity prediction, letting us ask a sharper question: with the action fixed, can video backbones adapt to identity-discriminative evidence in \textit{how} it is executed?

We use pre-trained AR backbones as a diagnostic basis. These models are not built for identity recognition, but their representations encode motion dynamics and body configuration. We repurpose them by replacing the action classification head with a linear identity classifier and fully fine-tuning the backbone (\eg, UniFormerV2~\cite{li2023uniformerv2} pre-trained on Kinetics-400~\cite{kay2017kinetics}). This does not assume the learned evidence is purely motion-based; rather, it lets us compare how identity evidence changes as appearance is available or suppressed. As shown in~\cref{fig:ar-to-identity}, the backbone maps each video to an embedding $\mathbf{e}$ optimized for identity, applied identically across appearance, silhouette, and skeleton regimes.

We train with two complementary objectives. The \textit{\textbf{classification path}} feeds $\mathbf{e}$ into the identity classifier to produce logits $\mathbf{z}$ over $N$ identities, with cross-entropy and label smoothing ($\epsilon=0.1$):
\begin{equation}
    \mathcal{L}_{id} = - \sum_{k=1}^{N} q(k) \log p(k),
\end{equation}
where $p(k)=\operatorname{softmax}(\mathbf{z})_{k}$ and $q(k)=(1-\epsilon)\delta_{k,y}+\epsilon/N$ for true label $y$. The \textit{\textbf{metric-learning path}} applies a batch-hard triplet loss to $\mathbf{e}$, pulling same-identity samples closer than different-identity ones by a margin $m=0.3$:
\begin{equation}
    \mathcal{L}_{tri}
    =
    \sum_{i=1}^{B}
    \left[
    \max_{\mathbf{e}_{p} \in \mathcal{P}_{i}}
    d(\mathbf{e}_{i}, \mathbf{e}_{p})
    -
    \min_{\mathbf{e}_{n} \in \mathcal{N}_{i}}
    d(\mathbf{e}_{i}, \mathbf{e}_{n})
    + m
    \right]_{+},
\end{equation}
where $B$ is the batch size, $[\cdot]_{+}=\max(0,\cdot)$, $d(\cdot,\cdot)$ is the Euclidean distance, and $\mathcal{P}_{i}$, $\mathcal{N}_{i}$ denote the positive and negative samples for anchor $\mathbf{e}_{i}$ within the batch. The final objective is
\begin{equation}
    \mathcal{L}_{total} = \mathcal{L}_{id} + \mathcal{L}_{tri}.
\end{equation}

\section{Experiment and Analysis}
\label{sec:exp and anal}

\subsection{Experimental Setup}
\label{sec:exp-setup}
\begin{table}[t]
\centering
\let\rawdiffup\diffup \let\rawdiffdown\diffdown \let\rawdiffsame\diffsame
\renewcommand{\diffup}[1]{\makebox[2.6em][l]{\rawdiffup{#1}}}
\renewcommand{\diffdown}[1]{\makebox[2.6em][l]{\rawdiffdown{#1}}}
\renewcommand{\diffsame}[1]{\makebox[2.6em][l]{\rawdiffsame{#1}}}
\caption{Closed-set performance under the \textbf{standard} split. For each backbone, the gap is computed relative to the appearance-input model; gait methods are evaluated as standalone baselines.}
\vspace{-8pt}
\label{tab:closed-set-standard}
\setlength{\tabcolsep}{2pt}
\renewcommand{\arraystretch}{1.05}
\begin{adjustbox}{max width=\columnwidth}
\begin{tabular}{l l l l l l l l}
\toprule
\multirow{2}{*}{\textbf{Model}} & \multirow{2}{*}{\textbf{Input}}
& \multicolumn{3}{c}{\textbf{Classification}} & \multicolumn{3}{c}{\textbf{Retrieval}} \\
\cmidrule(lr){3-5} \cmidrule(lr){6-8}
& & \textbf{Top-1 (\%)} & \textbf{Top-3 (\%)} & \textbf{Top-5 (\%)} & \textbf{mAP (\%)} & \textbf{R-1 (\%)} & \textbf{R-5 (\%)} \\
\midrule
\multirow{3}{*}{MViTv2~\cite{li2022mvitv2}}
& App.  & $98.60$ & $99.30$ & $99.51$ & $98.46$ & $98.82$ & $98.82$ \\
& \cs Sil.  
    & \cs $98.39$\,\diffdown{0.21} 
    & \cs $99.37$\,\diffup{0.07} 
    & \cs $99.51$\,\diffsame{0.00}
    & \cs $98.31$\,\diffdown{0.15} 
    & \cs $98.67$\,\diffdown{0.15} 
    & \cs $98.82$\,\diffsame{0.00} \\
& \ck Skel.\textsuperscript{$\clubsuit$} 
    & \ck $97.84$\,\diffdown{0.76} 
    & \ck $99.16$\,\diffdown{0.14} 
    & \ck $99.30$\,\diffdown{0.21}
    & \ck $97.07$\,\diffdown{1.39} 
    & \ck $97.64$\,\diffdown{1.18} 
    & \ck $97.79$\,\diffdown{1.03} \\
\midrule
\multirow{2}{*}{VideoMAEV2~\cite{wang2023videomae}}
& App.  & $98.04$ & $98.95$ & $99.16$ & $97.93$ & $98.53$ & $98.53$ \\
& \cs Sil.  
    & \cs $95.46$\,\diffdown{2.58} 
    & \cs $98.67$\,\diffdown{0.28} 
    & \cs $99.09$\,\diffdown{0.07}
    & \cs $95.58$\,\diffdown{2.35} 
    & \cs $97.64$\,\diffdown{0.89} 
    & \cs $98.08$\,\diffdown{0.45} \\
\midrule
\multirow{2}{*}{UniFormerV2~\cite{li2023uniformerv2}}
& App.  & $97.49$ & $99.02$ & $99.30$ & $89.35$ & $97.20$ & $98.97$ \\
& \cs Sil.  
    & \cs $95.67$\,\diffdown{1.82} 
    & \cs $98.25$\,\diffdown{0.77} 
    & \cs $98.95$\,\diffdown{0.35}
    & \cs $85.95$\,\diffdown{3.40} 
    & \cs $95.13$\,\diffdown{2.07} 
    & \cs $97.49$\,\diffdown{1.48} \\
\midrule
\rowcolor{blue!5}
\multicolumn{8}{c}{\textit{\textbf{Gait Recognition Methods}}} \\
\midrule
BigGait~\cite{ye2024biggait}
& App. & $99.51$ & $99.79$ & $99.86$ & $98.27$ & $99.41$ & $99.41$ \\
DeepGaitV2~\cite{fan2023exploring}
& Sil. & $99.72$ & $99.79$ & $99.79$ & $98.72$ & $99.26$ & $99.41$ \\
\bottomrule
\end{tabular}
\end{adjustbox}
\par\vspace{3pt}
\raggedright
{\footnotesize $^{\clubsuit}$\,We only report the MViTv2 results for the skeleton input (see~\cref{ss_main_result}).}
\vspace{-13pt}
\end{table}
\begin{table}[t]
\centering
\let\rawdiffup\diffup \let\rawdiffdown\diffdown \let\rawdiffsame\diffsame
\renewcommand{\diffup}[1]{\makebox[2.6em][l]{\rawdiffup{#1}}}
\renewcommand{\diffdown}[1]{\makebox[2.6em][l]{\rawdiffdown{#1}}}
\renewcommand{\diffsame}[1]{\makebox[2.6em][l]{\rawdiffsame{#1}}}
\caption{\small Closed-set performance under the \textbf{appearance-disjoint} split. Gaps are relative to the same backbone's appearance input.}
\vspace{-7pt}
\label{tab:closed-set-disjoint}
\setlength{\tabcolsep}{2pt}
\renewcommand{\arraystretch}{1.05}
\begin{adjustbox}{max width=\columnwidth}
\begin{tabular}{l l l l l l l l}
\toprule
\multirow{2}{*}{\textbf{Model}} & \multirow{2}{*}{\textbf{Input}}
& \multicolumn{3}{c}{\textbf{Classification}} & \multicolumn{3}{c}{\textbf{Retrieval}} \\
\cmidrule(lr){3-5} \cmidrule(lr){6-8}
& & \textbf{Top-1 (\%)} & \textbf{Top-3 (\%)} & \textbf{Top-5 (\%)} & \textbf{mAP (\%)} & \textbf{R-1 (\%)} & \textbf{R-5 (\%)} \\
\midrule
\multirow{3}{*}{MViTv2~\cite{li2022mvitv2}}
& App.  & $5.31$ & $13.62$ & $19.24$ & $57.63$ & $84.90$ & $92.26$ \\
& \cs Sil.  
    & \cs $89.07$\,\diffup{83.76} 
    & \cs $95.66$\,\diffup{82.04} 
    & \cs $96.95$\,\diffup{77.71}
    & \cs $86.12$\,\diffup{28.49} 
    & \cs $92.64$\,\diffup{7.74} 
    & \cs $95.94$\,\diffup{3.68} \\
& \ck Skel. 
    & \ck $93.95$\,\diffup{88.64} 
    & \ck $97.68$\,\diffup{84.06} 
    & \ck $98.47$\,\diffup{79.23}
    & \ck $92.08$\,\diffup{34.45} 
    & \ck $96.32$\,\diffup{11.42} 
    & \ck $97.97$\,\diffup{5.71} \\
\midrule
\multirow{2}{*}{VideoMAEV2~\cite{wang2023videomae}}
& App.  & $1.34$ & $6.54$ & $10.20$ & $59.61$ & $89.72$ & $94.04$ \\
& \cs Sil.  
    & \cs $84.42$\,\diffup{83.08} 
    & \cs $92.91$\,\diffup{86.37} 
    & \cs $95.42$\,\diffup{85.22}
    & \cs $78.94$\,\diffup{19.33} 
    & \cs $90.86$\,\diffup{1.14} 
    & \cs $94.92$\,\diffup{0.88} \\
\midrule
\multirow{2}{*}{UniFormerV2~\cite{li2023uniformerv2}}
& App.  & $22.85$ & $43.80$ & $57.06$ & $54.70$ & $85.66$ & $93.27$ \\
& \cs Sil.  
    & \cs $88.82$\,\diffup{65.97} 
    & \cs $94.56$\,\diffup{50.76} 
    & \cs $95.60$\,\diffup{38.54}
    & \cs $74.83$\,\diffup{20.13} 
    & \cs $92.89$\,\diffup{7.23} 
    & \cs $97.46$\,\diffup{4.19} \\
\bottomrule
\end{tabular}
\end{adjustbox}
% \par\vspace{3pt}
% \raggedright
% {\footnotesize $^{\clubsuit}$\,We report the skeleton modality only for MViTv2; the other two backbones are not compatible with this input. See the supplementary material for details.}
\vspace{-12pt}
\end{table}
\noindent\textbf{Model Architecture.}~We adapt three pre-trained action-recognition backbones for identity recognition: MViTv2~\cite{li2022mvitv2}, VideoMAEV2~\cite{wang2023videomae}, and UniFormerV2~\cite{li2023uniformerv2}. MViTv2 and UniFormerV2 are initialized with Kinetics-400 weights, while VideoMAEV2 uses Kinetics-710 weights. VideoMAEV2 and UniFormerV2 use ViT-B/16 encoders, whereas MViTv2 uses the small variant. We further compare against two dedicated motion-based identity recognition baselines, BigGait~\cite{ye2024biggait} and DeepGaitV2~\cite{fan2023exploring}.
 
% We adapt pre-trained action-recognition backbones for identity recognition. We consider three transformer architectures: MViTv2~\cite{li2022mvitv2}, VideoMAEV2 \cite{wang2023videomae}, and~UniFormerV2~\cite{li2023uniformerv2}, all initialized with Kinetics-400 pre-trained weights. VideoMAEV2 and UniFormerV2 use ViT-B/16 as the visual encoder, while MViTv2 uses its small variant. We additionally include BigGait~\cite{ye2024biggait} and DeepGaitV2~\cite{fan2023exploring} as dedicated motion-based identity baselines.

% \noindent\textbf{Input regimes.}
% We evaluate three input regimes, following the pipeline described in~\cref{sec:dataset}: appearance, silhouette, and skeleton. Appearance videos visualize the tracked player as masked RGB crops with background context removed. Silhouette videos visualize the same sequence as binary player masks, suppressing facial details, jersey texture, color, and logos. Skeleton videos visualize sparse body keypoints estimated from the appearance frames, retaining joint configurations and their temporal progression.

\noindent\textbf{Metric.} We report both the accuracy of the identity classifier and the retrieval performance of the learned feature embeddings (\eg, mAP and Rank-$k$: whether the top-$k$ retrieved clips contain at least one correct identity). 

\noindent\textbf{Data Split.} We use two strategies. The standard split divides each identity's clips into train/test at 7:3, with test clips further split into query/gallery at 5:5. The appearance-disjoint split instead partitions clips by jersey appearance, so each identity's train and test clips have disjoint appearance, using the same query/gallery protocol.

\noindent\textbf{Implementation Details.}
All models are trained for 100 epochs using AdamW~\cite{loshchilov2017decoupled} with cosine annealing. Input frames are resized to $224 \times 224$, and each clip is sampled to 16 frames. We use random horizontal flipping, color jittering, random erasing, and a random identity sampler during training, and uniform frame sampling during testing. We evaluate three input regimes, following the pipeline described in~\cref{sec:dataset}: appearance, silhouette, and skeleton.

\subsection{Main Results}
\label{ss_main_result}
% We organize the results around three questions: whether video models can recognize identity, whether this recognition survives appearance changes, and what evidence the models use.

\noindent\textbf{Can General Video Backbones Recognize Identity?}
General video backbones recognize players reliably once repurposed for identity, even though identity recognition is not their original design goal. As shown in~\Cref{tab:closed-set-standard}, under the standard split, where each identity's clips are divided randomly so that appearance distributions overlap between training and testing, full-appearance input drives all three action-recognition backbones to near-perfect accuracy.

A deeper question is what remains once appearance is removed. The silhouette and skeleton regimes strip away face, jersey, and texture, keeping how the body is configured and how it moves. Both stay competitive with full appearance, dropping only marginally. Identity-specific motion is therefore not merely present in these clips; it is readily learnable even when appearance is suppressed. The dedicated gait models reach comparable near-ceiling accuracy under the same protocol, suggesting that the identity signal in free-throw execution is not specific to a single architecture.

Interestingly, the two appearance-suppressed regimes are not equivalent. Silhouettes are more compatible with standard video backbones, likely because they remain video-like while preserving dense body shape and fine-grained motion cues that sparse skeletons may discard. They perform consistently well across architectures. Skeletons transfer less reliably; only MViTv2 remains robust in this regime, so we report skeleton results only for MViTv2.

%Interestingly, the two suppressed regimes are not equivalent. We find that silhouettes tend to be the more compatible modality: they appear to preserve finer, identity-unique motion that the sparse skeleton may collapse, and they remain effective across all backbones. Skeletons, by contrast, transfer less reliably---only MViTv2 stays robust under this regime, which is why we report its skeleton results alone. 
% \input{table_tex/closed-set-appearance}
% \input{table_tex/image-study}

\begin{figure}[b]
    \centering
    \vskip -10pt
    \includegraphics[width=0.95\linewidth]{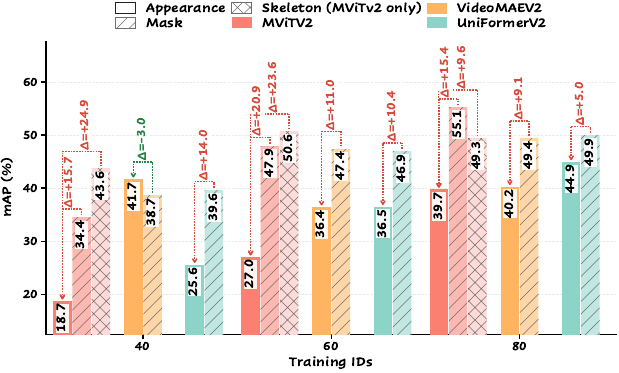}
    \vskip -10pt
    \caption{Open-set performance under the appearance-disjoint setting. Appearance-suppressed regimes (striped, diamond-hatched bars) stay mostly ahead across training scales, indicating that motion-based representations transfer to unseen identities.}
    \label{fig:open-set}
\end{figure}

%Different input regimes may carry different identity cues. Under the standard split, where appearance and motion both transfer, which cue the models rely on stays unclear. To separate the two, we move to the appearance-disjoint split. As shown in~\Cref{tab:closed-set-disjoint}, where each identity's training and test clips are separated by jersey appearance, full-appearance models degrade sharply, while the two appearance-suppressed regimes remain robust and outperform appearance across the board. Since full-appearance videos contain the same motion available to silhouettes and skeletons, this gap indicates that appearance models do not lack motion cues; they lean on appearance ones that fail to transfer once jersey changes. The cues recovered under suppression, by contrast, are not generic shortcuts but motion patterns specific to each individual.

\noindent\textbf{What Survives When Appearance Shifts?} Models trained under different regimes may rely on different identity cues. Under the standard split, where both appearance and motion can transfer, it remains unclear which cues drive recognition. To separate them, we evaluate the appearance-disjoint split. As shown in~\Cref{tab:closed-set-disjoint}, when each identity's training and test clips differ in jersey appearance, full-appearance models degrade sharply, while appearance-suppressed models remain robust and outperform appearance models across the board. Since full-appearance videos contain the same motion information available to silhouettes and skeletons, this gap indicates that appearance models do not fail because motion cues are absent; rather, they rely on appearance cues that do not transfer once jerseys change. By contrast, the cues recovered under suppression are not generic shortcuts, but individual-specific motion patterns.

We next test whether these motion cues extend to unseen identities, \ie, the open-set setting. Under the appearance-disjoint split, we hold out 40 test identities and gradually increase the number of training identities. As shown in~\cref{fig:open-set}, appearance-suppressed regimes perform the best mostly, with gains persisting as training scale grows. With appearance shortcuts shifted and identities unseen, recognition must rely more on subtle motion cues that appearance usually overshadows, indicating that these cues are genuine and transferable rather than tied to the training identities.

\begin{table}[t]
    \centering
    \caption{Impact of contour degradation ($\mathcal{B}$) on silhouette-based (Sil.) models. Despite the degradation, all models retain strong performance, confirming that recognition relies on motion dynamics rather than fine-grained boundary details.}
    \vspace{-10pt}
    \label{tab:contour_degradation}
    \let\rawdiffup\diffup \let\rawdiffdown\diffdown \let\rawdiffsame\diffsame
    \renewcommand{\diffup}[1]{\makebox[2.6em][l]{\rawdiffup{#1}}}
    \renewcommand{\diffdown}[1]{\makebox[2.6em][l]{\rawdiffdown{#1}}}
    \renewcommand{\diffsame}[1]{\makebox[2.6em][l]{\rawdiffsame{#1}}}
    \setlength{\tabcolsep}{2pt}
    \renewcommand{\arraystretch}{1.05}
    \begin{adjustbox}{max width=\columnwidth}
    \begin{tabular}{llllllll}
    \toprule
    \multirow{2}{*}{\textbf{Model}} &
    \multirow{2}{*}{\textbf{Input}} &
    \multicolumn{3}{c}{\textbf{Classification}} &
    \multicolumn{3}{c}{\textbf{Retrieval}} \\
    \cmidrule(lr){3-5} \cmidrule(lr){6-8}
    & & Top-1 (\%) & Top-3 (\%) & Top-5 (\%) & mAP (\%) & R-1 (\%) & R-5 (\%) \\
    \midrule
    \multirow{2}{*}{MViTv2~\cite{li2022mvitv2}}
    & Sil. & $98.39$ & $99.37$ & $99.51$ & $98.31$ & $98.67$ & $98.82$ \\
    & \cb Sil.$+\mathcal{B}$ 
        & \cb $97.00$\,\diffdown{1.39} 
        & \cb $99.16$\,\diffdown{0.21} 
        & \cb $99.37$\,\diffdown{0.14}
        & \cb $95.11$\,\diffdown{3.20} 
        & \cb $97.49$\,\diffdown{1.18} 
        & \cb $98.23$\,\diffdown{0.59} \\
    \midrule
    \multirow{2}{*}{VideoMAEV2~\cite{wang2023videomae}}
    & Sil. & $95.46$ & $98.67$ & $99.09$ & $95.58$ & $97.64$ & $98.08$ \\
    & \cb Sil.$+\mathcal{B}$ 
        & \cb $92.04$\,\diffdown{3.42} 
        & \cb $96.37$\,\diffdown{2.30} 
        & \cb $97.49$\,\diffdown{1.60}
        & \cb $88.38$\,\diffdown{7.20} 
        & \cb $92.77$\,\diffdown{4.87} 
        & \cb $95.28$\,\diffdown{2.80} \\
    \midrule
    \multirow{2}{*}{UniFormerV2~\cite{li2023uniformerv2}}
    & Sil. & $95.67$ & $98.25$ & $98.95$ & $85.95$ & $95.13$ & $97.49$ \\
    & \cb Sil.$+\mathcal{B}$ 
        & \cb $94.76$\,\diffdown{0.91} 
        & \cb $97.56$\,\diffdown{0.69} 
        & \cb $98.18$\,\diffdown{0.77}
        & \cb $83.29$\,\diffdown{2.66} 
        & \cb $94.10$\,\diffdown{1.03} 
        & \cb $97.64$\,\diffup{0.15} \\
    \bottomrule
    \end{tabular}
    \end{adjustbox}
    \vspace{-10pt}
\end{table}
\begin{figure}[t]
    \centering
    \includegraphics[width=\linewidth]{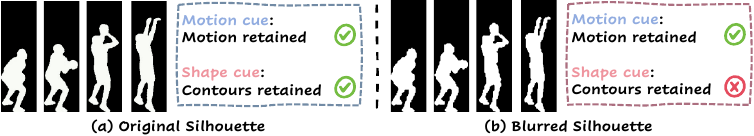}
    \vspace{-19pt}
    \caption{Original vs.\ contour-degraded silhouettes: fine boundaries are blurred while body motion is retained.}
    \label{fig:contour}
    \vspace{-15pt}
\end{figure}

\noindent\textbf{Sanity Check: Shape or Execution?}~Silhouettes suppress face, jersey, and texture, but their contours still trace a body outline, so recognition could rest on fixed shape rather than on how the action is executed. We test this by degrading contour fidelity while leaving the temporal structure intact. As illustrated in~\cref{fig:contour}, each binary mask is downsampled and reconstructed by bicubic interpolation, blurring fine boundaries while preserving how the body moves; the shape cue is weakened, the execution cue is kept. As shown in~\Cref{tab:contour_degradation}, recognition is largely unaffected: across all three backbones, the drop is marginal, and accuracy stays high even when boundaries are coarse. If the models were reading identity from precise contours, this degradation should substantially reduce performance. The observed stability suggests instead that recognition does not hinge on fixed shape, but on execution-dependent evidence.

\subsection{Main Analysis}
\label{sec:main-analysis}
To reveal these identity-specific \textit{motion signatures} in a verifiable, interpretable form, we use CAM-based saliency~\cite{zhou2016learning}, examining whether the evidence is tied to each player's execution, stable across views, and distinct across players.

\begin{figure}[h]
    \centering
    \includegraphics[width=0.95\linewidth]{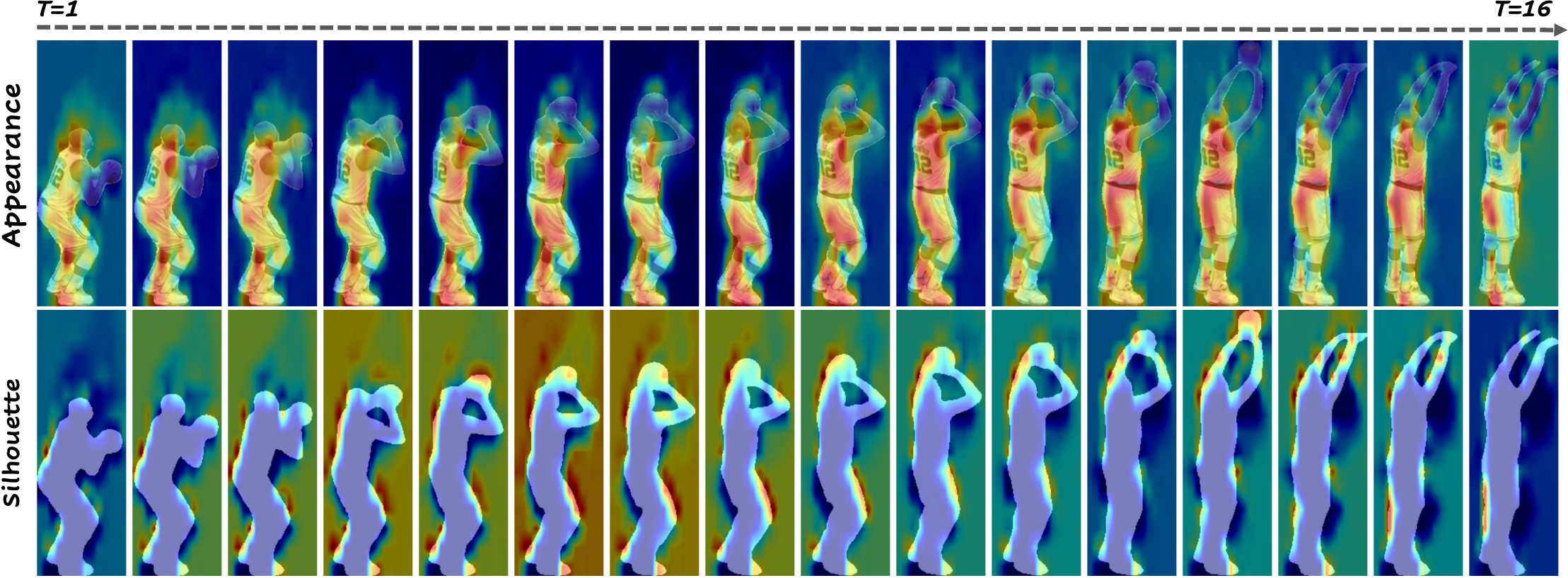}
    \vspace{-7pt}
    \caption{Saliency maps for the same player (Al Horford). The appearance model attends to jersey and face regions, whereas the silhouette model shifts with the unfolding free-throw.}
    \label{fig:exp1-app-vs-sil}
    \vspace{-20pt}
\end{figure}

\begin{figure}[b]
    \centering
    \vskip -15pt
    \includegraphics[width=\linewidth]{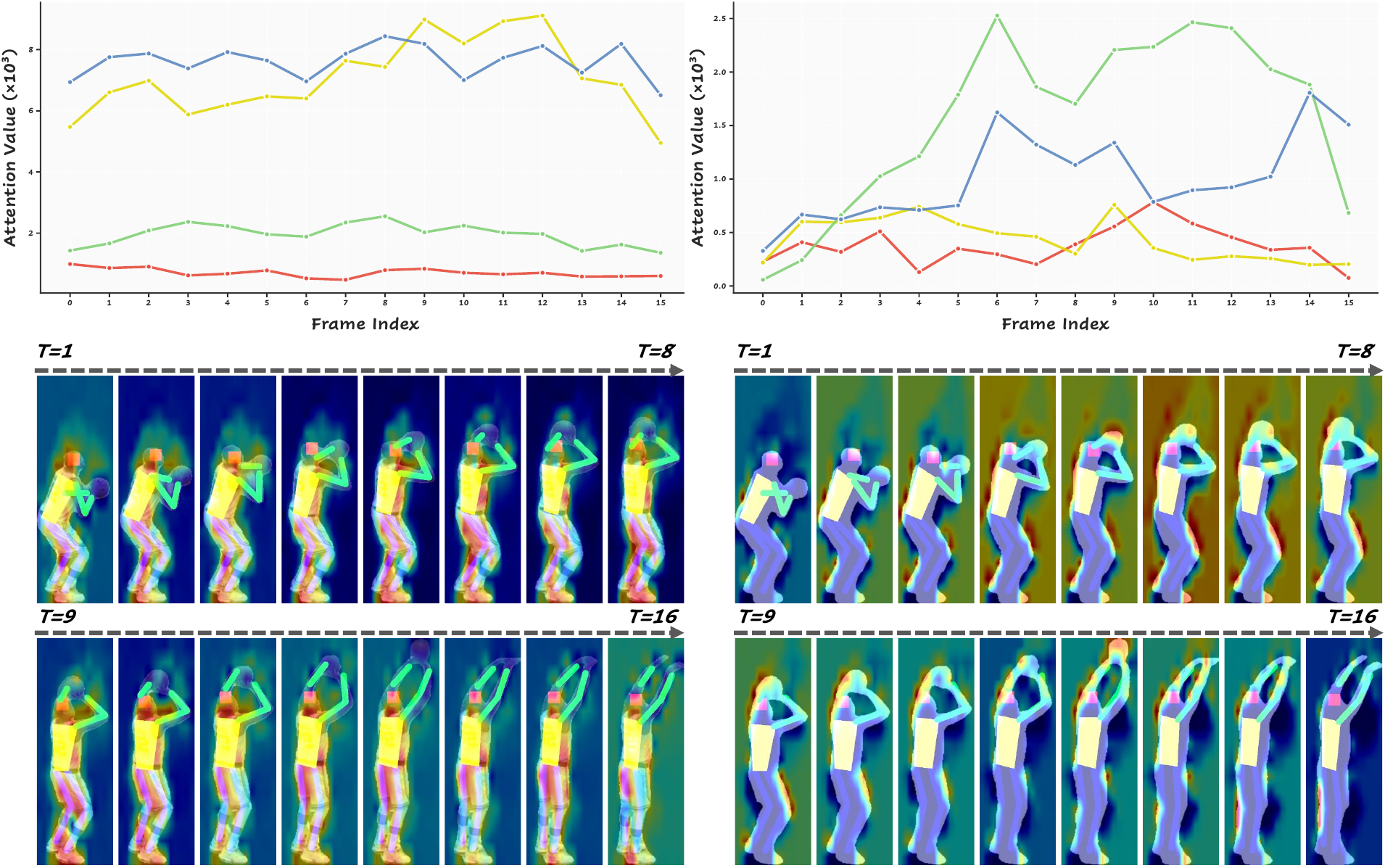}
    \vskip -10pt
    \caption{Pose-based quantification of attention dynamics. Saliency is aggregated per frame over four regions---\textcolor{red}{Head}, \textcolor[HTML]{E2D909}{Torso}, \textcolor[HTML]{82D177}{Arms}, \textcolor[HTML]{5E89C2}{Legs}. Appearance saliency is flat over time; silhouette saliency tracks the shot phases.}
    % \vskip -10pt
    \label{fig:exp2-quantitative-app-vs-sil}
\end{figure}

\noindent\textbf{Different Regimes Attend to Different Cues.}
As shown in~\cref{fig:exp1-app-vs-sil}, the two regimes focus on different evidence for the same player. The appearance model focuses on the jersey and face, holding attention on static regions across the whole sequence. The silhouette model instead shifts its focus as the action unfolds: attention starts on the back and legs as the shot begins, shifts to the thighs and arms during the rise, and concentrates on the arms at release. This frame-by-frame alignment with the shooting motion indicates that the silhouette model tracks how the action is executed rather than what the player looks like.
%\noindent\textbf{Attention moves with the motion.}
Beyond visual inspection, we quantify saliency by grouping pixels into four regions---head, torso, arms, and legs---and aggregating attention within each region per frame (\cref{fig:exp2-quantitative-app-vs-sil}). For appearance input, regional saliency stays flat across the sequence, dominated by the jersey-bearing torso and legs. For silhouette input, the curves rise and fall with the phases of the shot: arm and leg saliency grows as the limbs elevate, while the torso stays low throughout. The attention is therefore not a static spatial bias but a temporally organized response that follows the execution itself.

\begin{figure}[t]
    \centering
    \includegraphics[width=0.9\linewidth]{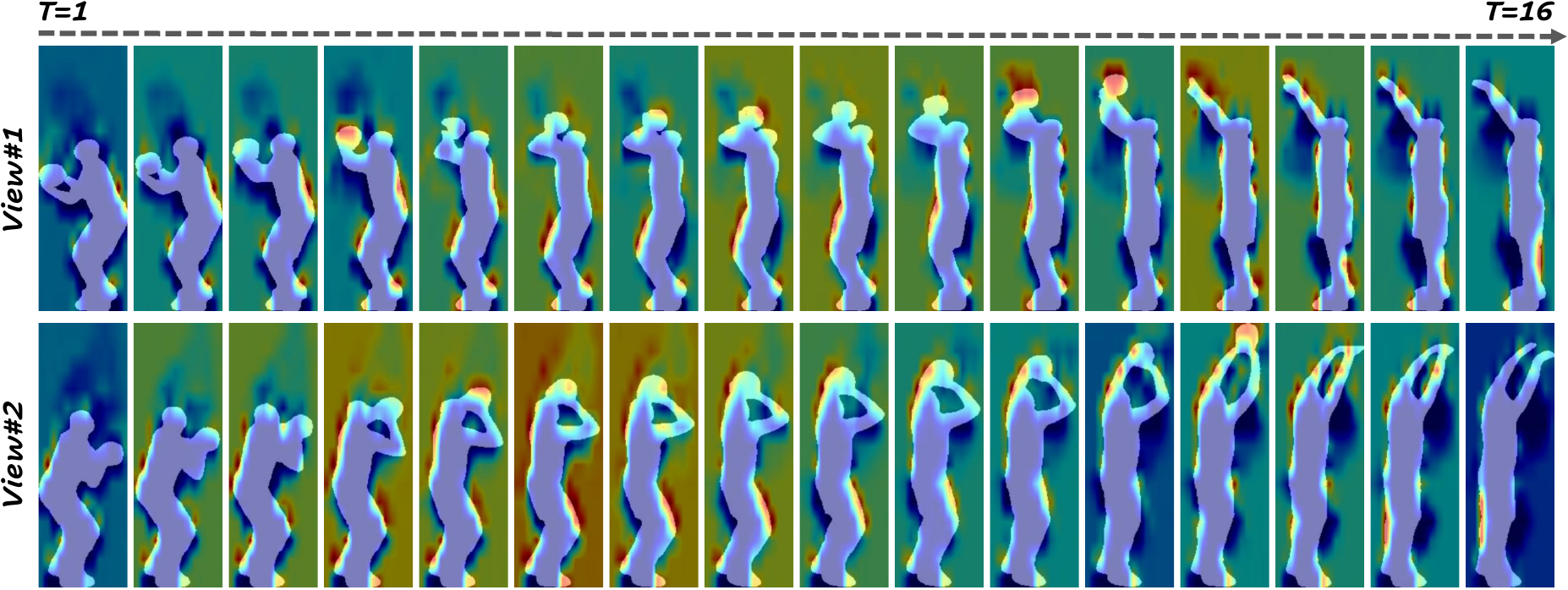}
    \vspace{-8pt}
    \caption{Saliency maps of the same player across different camera viewpoints. The silhouette model follows the same body regions in the same temporal order, showing that the motion evidence is stable within an identity.}
    \label{fig:exp3-cross_scene_consistency}
    \vspace{-10pt}
\end{figure}

\noindent\textbf{Signature Holds Across Viewpoints.} Motion signatures should persist regardless of viewpoint. ~In~\cref{fig:exp3-cross_scene_consistency}, we compare saliency for the same player across clips captured from different broadcast angles. Despite the change in viewpoint, the silhouette model traverses the same body regions in the same temporal order, reproducing the player's execution pattern rather than a view-dependent artifact. The evidence the model relies on is thus stable within an identity.

\begin{figure}[h]
    \centering
    \vskip-10pt
    \includegraphics[width=0.95\linewidth]{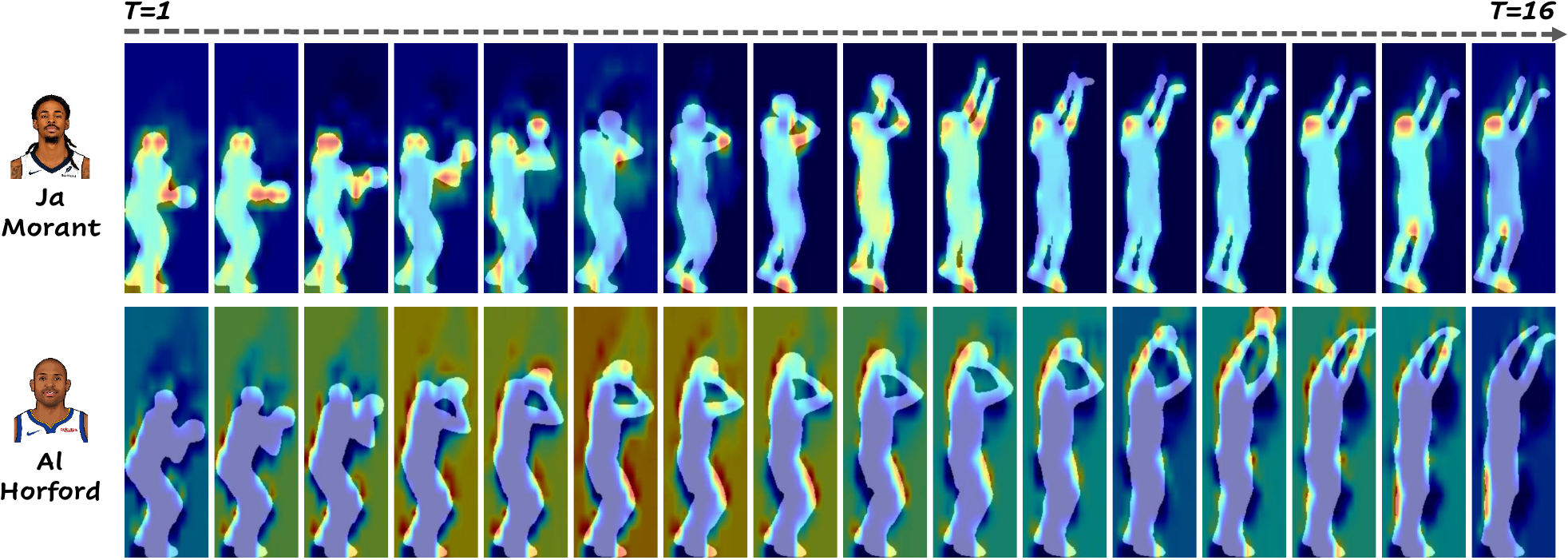}
    \vspace{-8pt}
    \caption{Saliency maps for Ja Morant and Al Horford performing free-throws. The model attends to different body regions per player, showing that the evidence is distinct across identities.}
    \label{fig:exp4-different-player}
    \vspace{-7pt}
\end{figure}

\noindent\textbf{Each Player Has a Distinctive Signature.}
%The same evidence also separates players. 
In~\cref{fig:exp4-different-player}, two players performing the identical action draw distinct attention. For the fast, explosive shooter, the model concentrates on rapid arm elevation and elbow trajectory in the early frames, whereas for the more deliberate shooter, it attends to foot placement and arm configuration throughout, including the post-release follow-through. Applied to the same action, the model surfaces different evidence for each identity---confirming that what it captures reflects individual execution style, not a generic free-throw template.

\subsection{Further Exploration and Investigation }
\label{sec:further-exploration}
\begin{figure}[t]
  \centering
  \includegraphics[width=0.9\linewidth]{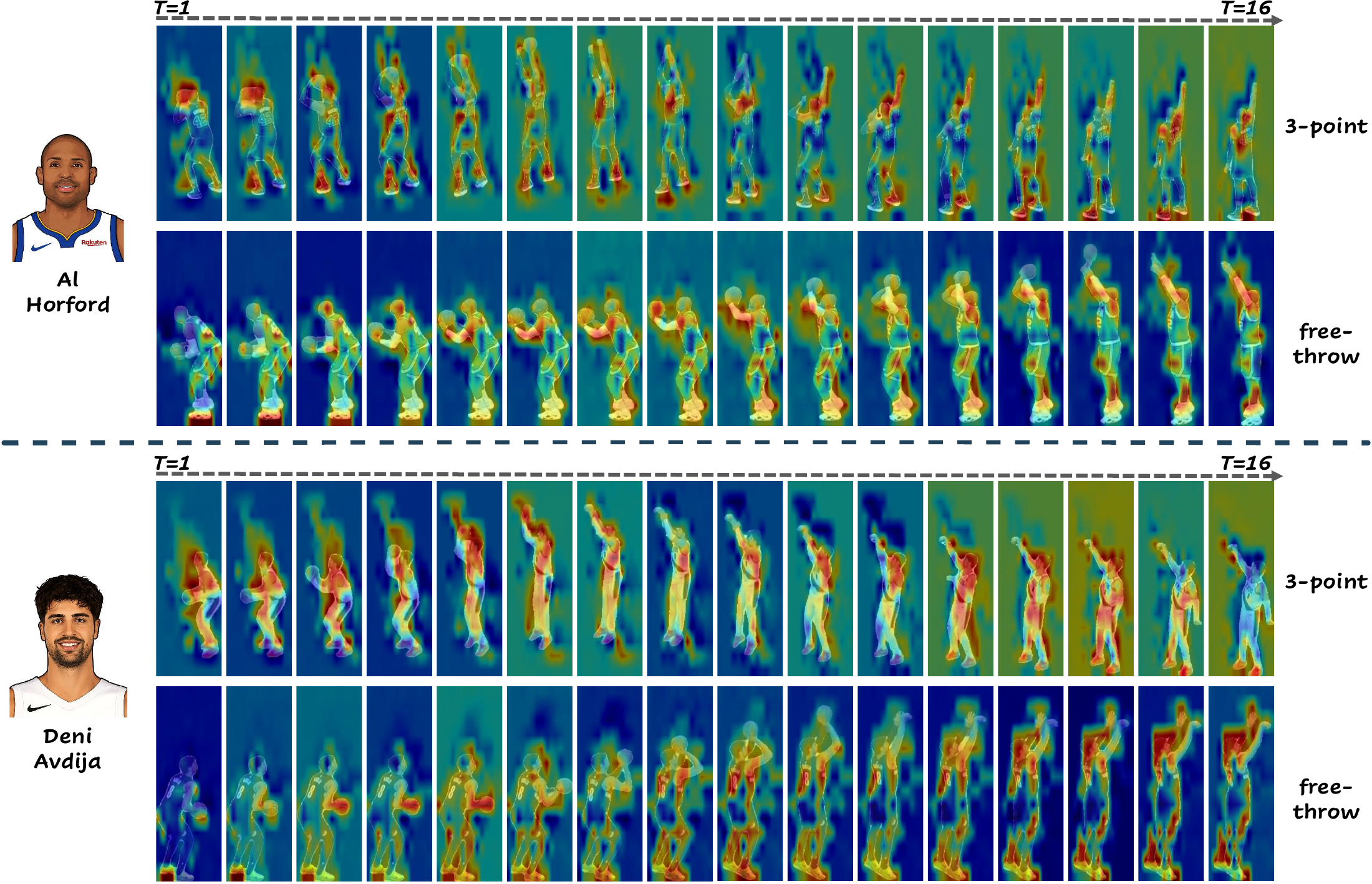}
  \vspace{-8pt}
    \caption{Saliency maps under action recognition between three-pointers and free-throws. Identity-specific motions are overshadowed by shared, action-defining patterns common across players.}
  \label{fig:3pt_vs_freethrow}
  \vspace{-5pt}
\end{figure}
\noindent\textbf{Is Appearance Suppression Needed to Use Motion Cues?} Silhouette models reliably encode identity-specific motion, whereas full-appearance models often do not. This raises a question: do video models use motion only when appearance is suppressed? To test this, we curate three-pointer clips for 109 players and fine-tune a full-appearance backbone for binary action recognition: three-pointers vs.~free-throws. As shown in~\cref{fig:3pt_vs_freethrow}, the task changes what the model attends to. For action recognition, saliency follows the shooting execution---arm elevation, release trajectory, and torso lift---even without appearance suppression. Yet these cues are shared across players: they capture \emph{what} action is performed, not \emph{who} performs it. For identity recognition, the same model family shifts toward static appearance, suggesting that appearance provides an easier identity signal that overshadows subtler kinematic cues. We call this \textit{task-induced appearance bias}: models favor visual shortcuts when they predict the training objective. Only by fixing the action and suppressing appearance are models pushed toward genuine identity-specific motion. This further supports our controlled design: unconstrained actions introduce semantic variation, while intact appearance draws models toward shortcuts and away from the execution cues needed to study motion-based identity recognition.

%\noindent{\textbf{When Do Motion Signatures Emerge?}} Silhouette models reliably encode identity-specific motion, but does this arise on its own, or only under identity supervision paired with appearance suppression? As motivated in~\cref{sec:protocol}, high-variance actions hand models easy semantic shortcuts. To test this, we add three-point clips for 109 players, forming a balanced two-class action setting.~\cref{fig:3pt_vs_freethrow} shows how the objective reshapes what the model attends to. Under an action objective, saliency tracks the shooting execution---arm elevation, release trajectory, torso lift---and responds far more strongly to the explosive three-pointer than to the subtler free-throw. Yet this attention is shared across players: it captures what action is performed, not who performs it. An identity objective reverses the picture: saliency drops the shared action motion for static textures, since appearance offers an immediate identity signal that overshadows subtle kinematics. We call this \textit{task-induced appearance bias}---models favor visual shortcuts over the motion that distinguishes individuals. Only by fixing the action and suppressing appearance can the models be pushed toward genuine, identity-specific motion. This is why our controlled setting is a scientific choice, not a limitation: unconstrained actions add semantic variance that drowns out personal execution, and intact appearance triggers the bias above.
\begin{table}[t]
\centering
\caption{Probing whether static frames carry motion signatures, across four input regimes: App., Sil., Sil.$+\mathcal{B}$, and Skel. %Values are mean$\pm$std.
}
\vspace{-10pt}
\label{tab:img-study}
\setlength{\tabcolsep}{4pt}
\renewcommand{\arraystretch}{1.15}
\footnotesize
\begin{adjustbox}{max width=\linewidth}
\newcommand{\std}[1]{_{\textcolor{blue}{\pm#1}}}
\begin{tabular}{l c ll ll ll ll}
\toprule
\multirow{2}{*}{\textbf{Input}} 
& \multirow{2}{*}{\textbf{Frame(s)}}
& \multicolumn{2}{c}{\textbf{App.}}
& \multicolumn{2}{c}{\cellcolor{rowsil}\textbf{Sil.}}
& \multicolumn{2}{c}{\cellcolor{rowblur}\textbf{Sil.$+\mathcal{B}$}}
& \multicolumn{2}{c}{\cellcolor{rowskel}\textbf{Skel.}} \\
\cmidrule(lr){3-4}\cmidrule(lr){5-6}\cmidrule(lr){7-8}\cmidrule(lr){9-10}
& & \multicolumn{1}{c}{Acc (\%)} & \multicolumn{1}{c}{Conf}
  & \multicolumn{1}{c}{Acc (\%)} & \multicolumn{1}{c}{Conf}
  & \multicolumn{1}{c}{Acc (\%)} & \multicolumn{1}{c}{Conf}
  & \multicolumn{1}{c}{Acc (\%)} & \multicolumn{1}{c}{Conf} \\
\midrule
Image (DINOv3~\cite{simeoni2025dinov3}) & First
  & $93.09\std{9.08}$ & $0.63\std{0.27}$
  & $77.23\std{18.10}$ & $0.49\std{0.29}$
  & $72.63\std{19.16}$ & $0.47\std{0.28}$
  & $45.60\std{19.74}$ & $0.36\std{0.24}$ \\
Image (DINOv3~\cite{simeoni2025dinov3}) & Multiple
  & $96.65\std{6.27}$ & $0.85\std{0.17}$
  & $90.43\std{11.19}$ & $0.69\std{0.24}$
  & $91.97\std{10.67}$ & $0.71\std{0.24}$
  & $93.92\std{9.03}$ & $0.65\std{0.24}$ \\
\midrule
Video (MViTv2~\cite{li2022mvitv2}) & Multiple
  & $98.60\std{4.46}$ & $0.99\std{0.05}$
  & $98.39\std{4.86}$ & $0.98\std{0.07}$
  & $97.00\std{5.86}$ & $0.96\std{0.12}$
  & $97.84\std{5.31}$ & $0.98\std{0.08}$ \\
Video (VideoMAEv2~\cite{wang2023videomae}) & Multiple
  & $98.04\std{5.31}$ & $1.00\std{0.03}$
  & $95.46\std{7.67}$ & $0.98\std{0.08}$
  & $92.04\std{12.01}$ & $0.96\std{0.13}$
  & \textemdash & \textemdash \\  
Video (UniFormerV2~\cite{li2023uniformerv2}) & Multiple
  & $97.49\std{5.43}$ & $0.83\std{0.22}$
  & $95.67\std{6.88}$ & $0.82\std{0.22}$
  & $94.76\std{7.80}$ & $0.79\std{0.24}$
  & \textemdash & \textemdash \\
\bottomrule
\end{tabular}
\end{adjustbox}
\vspace{-15pt}
\end{table}
\noindent{\textbf{Do Motion Signatures Reside in Static Frames?}} As discussed in~\cref{sec:protocol}, some identity cues may appear within individual phases, such as the set posture and release pose. We ask whether such within-phase poses are sufficient to identify a player without continuous video. To test this, we use an image backbone, DINOv3~\cite{simeoni2025dinov3}, with the same classification head and training protocol as before. The model is trained either on a single starting frame or on the same frames used by the video models, treated as independent still images. \Cref{tab:img-study} shows that static poses are informative: a single frame already supports identity recognition, and using all frames recovers much of the gap to video models. However, image models remain less confident than video models at comparable accuracy, with the key difference being whether frames are processed in temporal order. Static poses thus capture part of the motion signature, but temporal relationships across phases provide more robust identity evidence. Continuous motion, rather than isolated snapshots, remains the stronger cue.

% As premised in~\cref{sec:protocol}, some cues appear within a single phase---the set posture, release pose, or follow-through configuration---so we ask whether such within-phase positions are already enough to identify a player without continuous video. We probe this with an image backbone: DINOv3~\cite{simeoni2025dinov3} with the same classification head and training protocol as before, fed either a single starting frame or the same frames the video models receive, treated as independent stills.~\Cref{tab:img-study} shows that they are, to a point. A single frame already identifies players, and using the full set of frames recovers most of the gap to the video models, confirming that static poses do carry identity cues. By comparison, however, the image models stay markedly less confident than their video counterparts at matching accuracy---the only difference being whether the frames are read in temporal order. Static poses thus capture part of the signature, but the temporal relationship among phases is what makes the evidence robust---continuous motion, not isolated snapshots, remains the stronger identity cue.
\section{Conclusion}
\label{sec:conclusion}
We introduce~\ours, a controlled diagnostic probe, to study a fundamental question: when identity-specific motion is clearly present, do modern video models rely on it for recognition? When appearance is suppressed, video models under silhouette and skeleton regimes remain competitive in recognition; attribution further indicates that their evidence is tied to how each player executes the action. Yet without suppression, models still default to appearance shortcuts. These findings show that identity-specific motion signatures are present, learnable, and verifiable, but easily overlooked, pointing toward video models that more reliably capture fine-grained individual dynamics.

% We investigate whether modern video models, when identity-specific motion is clearly present, rely on it for recognition.~To this end, we introduce~\ours, a controlled diagnostic probe over a multi-phase skilled action that separates execution-independent appearance from execution-dependent evidence, isolating what models actually use.~Repurposed video backbones recognize identity reliably, and once appearance is suppressed, the silhouette and skeleton regimes remain competitive, stay robust under appearance shifts, and transfer to unseen identities. Attribution further indicates that this evidence is tied to how each player executes the action, stable within an identity, and distinct across identities. Models nonetheless default to appearance shortcuts unless suppression forces otherwise.~These findings suggest that identity-specific motion signatures are present, learnable, and verifiable, yet easily overlooked, pointing toward video models that more reliably capture fine-grained individual dynamics.

\section*{Acknowledgments}
This research is supported in part by grants from the National Science Foundation (Imageomics Institute: NSF OAC-2118240). We are grateful for the generous support from the Ohio Supercomputer Center.

{
    \small
    \bibliographystyle{ieeenat_fullname}
    \bibliography{main}
}
\clearpage
\appendix
\maketitlesupplementary
\setcounter{page}{1}
This supplementary material provides additional implementation details and analyses that support the diagnostic findings in the main paper. It is organized as follows:
\begin{itemize}
    \item Appendix~\ref{supp:training-details} reports the backbone configurations and training hyperparameters used in our experiments.
    \item Appendix~\ref{supp:saliency} clarifies how saliency maps are used as diagnostic evidence and defines our interpretation criteria.
    \item Appendix~\ref{supp:skeleton-standard} provides additional skeleton-input results and discusses why skeleton compatibility differs across backbones.
    \item Appendix~\ref{supp:skeleton-attribution} visualizes skeleton-based saliency patterns to inspect identity-specific execution evidence.
    \item Appendix~\ref{supp:standard-open-set} reports complementary open-set results under the standard split.
    \item Appendix~\ref{supp:static-temporal} examines whether identity evidence comes from static frames or ordered temporal progression.
    \item Appendix~\ref{supp:representation-evidence} provides representation evidence by comparing retrieval neighborhoods and embedding spaces across regimes.
    \item Appendix~\ref{supp:signature-stability} provides qualitative and quantitative evidence for stable and distinctive execution cues.
    \item Appendix~\ref{supp:backbone-saliency} compares saliency patterns across backbones to show architecture-dependent evidence.
\end{itemize}

\section{Additional Training Details}
\label{supp:training-details}
\begin{table}[b]
\renewcommand{\thetable}{A1}
\centering
\caption{Configuration of MViTv2, compared with the original configuration. Hyperparameter names follow the original configuration file. $\dagger$ denotes a value modified from the original setting.}
\label{tab:mvitv2}
\small
\setlength{\tabcolsep}{6pt}
\begin{adjustbox}{max width=\columnwidth}
\begin{tabular}{lll}
\toprule
\textbf{Parameter} & \textbf{Ours} & \textbf{Original} \\
\midrule
\texttt{BACKBONE}                   & \texttt{MViTv2-S} & \texttt{MViTv2-S} \\
\texttt{EMBED\_DIM}                 & $96$ & $96$ \\
\texttt{DEPTH}                      & $16$ & $16$ \\
\texttt{NUM\_HEADS}                 & $1$ & $1$ \\
\texttt{MLP\_RATIO}                 & $4.0$ & $4.0$ \\
\texttt{PATCH\_KERNEL}              & $(3, 7, 7)$ & $(3, 7, 7)$ \\
\texttt{PATCH\_STRIDE}              & $(2, 4, 4)$ & $(2, 4, 4)$ \\
\texttt{PATCH\_PADDING}             & $(1, 3, 3)$ & $(1, 3, 3)$ \\
\texttt{DIM\_MUL}                   & $\{1,3,14\}\times2$ & $\{1,3,14\}\times2$ \\
\texttt{HEAD\_MUL}                  & $\{1,3,14\}\times2$ & $\{1,3,14\}\times2$ \\
\texttt{POOL\_Q\_STRIDE}            & $(1,2,2)\ \text{at}\ \{1,3,14\}$ & $(1,2,2)\ \text{at}\ \{1,3,14\}$ \\
\texttt{POOL\_KV\_STRIDE\_ADAPTIVE} & $[1, 8, 8]$ & $[1, 8, 8]$ \\
\texttt{POOL\_KVQ\_KERNEL}          & $[3, 3, 3]$ & $[3, 3, 3]$ \\
\texttt{CLS\_EMBED\_ON}             & True & True \\
\texttt{DROPPATH\_RATE}             & $0.2$ & $0.2$ \\
\texttt{DROPOUT\_RATE}              & $0.0$ & $0.0$ \\
\bottomrule
\end{tabular}
\end{adjustbox}
\end{table}
\begin{table}[t]
\renewcommand{\thetable}{A2}
\centering
\caption{Configuration of VideoMAEv2, compared with the original configuration. Hyperparameter names follow the original configuration file. $\dagger$ denotes a value modified from the original setting.}
\label{tab:videomaev2}
\small
\setlength{\tabcolsep}{6pt}
\begin{adjustbox}{max width=\columnwidth}
\begin{tabular}{lll}
\toprule
\textbf{Parameter} & \textbf{Ours} & \textbf{Original} \\
\midrule
\texttt{BACKBONE}                     & \texttt{vit\_base\_patch16\_224} & \texttt{vit\_base\_patch16\_224} \\
\texttt{embed\_dim}                   & $768$ & $768$ \\
\texttt{depth}                        & $12$ & $12$ \\
\texttt{num\_heads}                   & $12$ & $12$ \\
\texttt{mlp\_ratio}                   & $4.0$ & $4.0$ \\
\texttt{patch\_size}                  & $16$ & $16$ \\
\texttt{tubelet\_size}                & $2$ & $2$ \\
\texttt{use\_mean\_pooling}           & True & True \\
\texttt{cos\_attn}                    & False & False \\
\texttt{drop\_path\_rate}$^{\dagger}$ & $0.0$ & $0.1$ \\
\texttt{drop\_rate}$^{\dagger}$       & $0.2$ & $0.0$ \\
\texttt{attn\_drop\_rate}             & $0.0$ & $0.0$ \\
\texttt{head\_drop\_rate}             & $0.0$ & $0.0$ \\
\bottomrule
\end{tabular}
\end{adjustbox}
\end{table}
\begin{table}[t]
\renewcommand{\thetable}{A3}
\centering
\caption{Configuration of UniFormerV2, compared with the original configuration. Hyperparameter names follow the original configuration file. $\dagger$ denotes a value modified from the original setting.}
\label{tab:uniformerv2}
\small
\setlength{\tabcolsep}{6pt}
\begin{adjustbox}{max width=\columnwidth}
\begin{tabular}{lll}
\toprule
\textbf{Parameter} & \textbf{Ours} & \textbf{Original} \\
\midrule
\texttt{BACKBONE}                   & \texttt{uniformerv2\_b16} & \texttt{uniformerv2\_b16} \\
\texttt{N\_DIM}                     & $768$ & $768$ \\
\texttt{N\_HEAD}                    & $12$ & $12$ \\
\texttt{N\_LAYERS}                  & $4$ & $4$ \\
\texttt{MLP\_FACTOR}                & $4.0$ & $4.0$ \\
\texttt{RETURN\_LIST}               & $[8, 9, 10, 11]$ & $[8, 9, 10, 11]$ \\
\texttt{TEMPORAL\_DOWNSAMPLE}       & False & False \\
\texttt{BACKBONE\_DROP\_PATH\_RATE} & $0.0$ & $0.0$ \\
\texttt{DROP\_PATH\_RATE}           & $0.0$ & $0.0$ \\
\texttt{MLP\_DROPOUT}               & $[0.5, 0.5, 0.5, 0.5]$ & $[0.5, 0.5, 0.5, 0.5]$ \\
\texttt{CLS\_DROPOUT}               & $0.5$ & $0.5$ \\
\texttt{NO\_LMHRA}$^{\dagger}$      & False & True \\
\texttt{DOUBLE\_LMHRA}              & True & True \\
\texttt{DW\_REDUCTION}              & $1.5$ & $1.5$ \\
\bottomrule
\end{tabular}
\end{adjustbox}
\end{table}
\label{supp:add-training-detail}
\subsection{Backbone Configurations}
\label{supp:backbone}
For all three video backbones, we largely inherit the configurations from the corresponding pre-trained action recognition models, modifying only a few hyperparameters that significantly influence the fine-tuning performance (see \Cref{tab:mvitv2,tab:videomaev2,tab:uniformerv2}). We adjust these hyperparameters using the closed-set standard split in \cref{sec:exp-setup}, and then fix them for all other experiments. Namely, we treat the test set of the standard split as the validation set. The test sets of other experiments are never used for hyperparameter selection. 

For ease of comparison, the hyperparameters in~\Cref{tab:mvitv2,tab:videomaev2,tab:uniformerv2} follow the descriptions used in the original configuration files/model definitions, and hyperparameters that differ from the original setting are marked with $\dagger$.

%For all three video backbones, we largely inherit the configuration of the corresponding pre-trained action recognition model and modify only a small number of hyperparameters. These modified values were selected after experimenting with several fine-tuning recipes, choosing the configuration that performs best on the test set of the standard split described in~\cref{sec:exp-setup}.~We emphasize that the test set was never exposed during this recipe search; it was used solely for final evaluation. For ease of comparison, the hyperparameters in~\Cref{tab:mvitv2,tab:videomaev2,tab:uniformerv2} follow the descriptions used in the original configuration files/model definitions, and hyperparameters that differ from the original setting are marked with $\dagger$.

We refer the reader to the official repositories for the original configurations that our settings are based on:
\begin{itemize}
  \item MViTv2: \faGithub~\href{https://github.com/facebookresearch/SlowFast/blob/main/configs/Kinetics/MVITv2_S_16x4.yaml}{facebookresearch/SlowFast}
  \item VideoMAEv2: \faGithub~\href{https://github.com/OpenGVLab/VideoMAEv2/blob/master/scripts/finetune/vit_b_k400_ft.sh}{OpenGVLab/VideoMAEv2}
  \item UniFormerV2: \faGithub~\href{https://github.com/OpenGVLab/UniFormerV2/blob/main/exp/k400/k400+k710_b16_f8x224/config.yaml}{OpenGVLab/UniFormerV2}
\end{itemize}

\subsection{Training Hyperparameters}
\begin{table}[t]
\renewcommand{\thetable}{A4}
\centering
\caption{Training hyperparameters for fine-tuning the three backbones.}
\label{tab:training-params}
\small
\setlength{\tabcolsep}{8pt}
\begin{adjustbox}{max width=\columnwidth}
\begin{tabular}{lccc}
\toprule
\textbf{Parameter} & \textbf{MViTv2} & \textbf{UniFormerV2} & \textbf{VideoMAEv2} \\
\midrule
Optimizer          & AdamW & AdamW & AdamW \\
Weight decay       & $0.05$ & $0.05$ & $0.05$ \\
LR schedule        & Cosine & Cosine & Cosine \\
Base (peak) LR     & $1\mathrm{e}{-4}$ & $1\mathrm{e}{-4}$ & $1\mathrm{e}{-4}$ \\
Final LR           & $1\mathrm{e}{-6}$ & $1\mathrm{e}{-6}$ & $1\mathrm{e}{-6}$ \\
Total epochs       & $100$ & $100$ & $100$ \\
Batch size         & $16$ & $16$ & $8$ \\
\bottomrule
\end{tabular}
\end{adjustbox}
\end{table}
\label{supp:training-hparams}
We summarize the optimization settings used to fine-tune the three backbones in Table~\ref{tab:training-params}. Some of these details were already stated in \cref{sec:exp-setup} of the main paper, and are repeated here for completeness. All three backbones are fully fine-tuned with AdamW (weight decay $0.05$), a peak learning rate of $1\mathrm{e}{-4}$, a cosine annealing learning-rate schedule decaying to a final learning rate of $1\mathrm{e}{-6}$, and a total of $100$ epochs. The only difference is the batch size: VideoMAEv2 uses a batch size of $8$ due to its higher memory footprint, while MViTv2 and UniFormerV2 use $16$.  

\section{Attribution Methods and Diagnostic Scope}
\label{supp:saliency}

\subsection{Background}
% \noindent{\textbf{Background.}}
Saliency-based attribution estimates which spatial or spatio-temporal regions of an input contribute to a trained model's prediction.~Existing approaches include perturbation- or optimization-based explanations~\cite{ribeiro2016should,lundberg2017unified}, gradient-based saliency~\cite{baehrens2010explain}, and class activation mapping (CAM) methods~\cite{zhou2016learning,selvaraju2017grad,chattopadhay2018grad,wang2020score,zhang2025finer}.~CAM-based methods are commonly used in visual recognition because they project class-discriminative evidence from intermediate feature maps back to the input domain. Grad-CAM~\cite{selvaraju2017grad} uses class-specific gradients as activation weights, while Score-CAM~\cite{wang2020score} estimates these weights through forward activation scores. For fine-grained recognition, Finer-CAM~\cite{zhang2025finer} further suppresses activations shared across visually similar classes to highlight more class-specific evidence.

\subsection{Role In Diagnostic Protocol}
% \noindent{\textbf{Role in our diagnostic protocol.}}
In~\ours, saliency maps are used as diagnostic support rather than standalone proof of identity-specific motion, following the diagnostic protocol described in~\cref{sec:protocol}. This distinction is important because appearance-suppressed inputs, especially silhouettes, still preserve body outline, phase-specific pose, and coarse configuration. We therefore do not interpret a highlighted silhouette region as direct evidence of ``pure motion.'' Instead, we use saliency to inspect how the evidence used by a trained identity classifier changes across the input regimes defined in~\cref{sec:dataset}: appearance, silhouette, and skeleton. Full-appearance models reveal whether the classifier relies on static visual shortcuts such as face, jersey texture, color, or number. Silhouette and skeleton models reveal whether the same architectures shift toward execution-dependent evidence, such as foot placement, elbow bend, torso configuration, limb coordination, or transitions across free-throw phases.

\subsection{Interpretation Criteria}
\label{supp:inter-criteria}
% \noindent{\textbf{Interpretation criteria.}}
Consistent with the analysis protocol in~\cref{sec:main-analysis}, we treat an attributed region as candidate identity-specific execution evidence only when it satisfies three criteria: (\textit{i}) it is stable across clips, games, and viewpoints for the same player; (\textit{ii}) it differs across players performing the same free-throw phase; and (\textit{iii}) it aligns with an observable component of the free-throw routine, such as set posture, rise, release, or follow-through. This phase-level reading is enabled by the controlled structure of~\ours. Accordingly, our saliency analysis is intended to localize candidate motion micro-signatures, while the overall claim is supported by the broader diagnostic design rather than by attribution alone. 

\section{Backbone Sensitivity to Skeleton Input}
\label{supp:skeleton-standard}

To complement the MViTv2 skeleton results discussed in~\cref{ss_main_result}, we additionally report the standard-split skeleton performance of VideoMAEv2 and UniFormerV2 in~\Cref{tab:skeleton-standard-backbones}. These results show that compatibility with skeleton input is strongly backbone-dependent.

\begin{table}[t]
\renewcommand{\thetable}{C1}
\centering
\caption{Skeleton-input performance under the \textbf{standard} split across three video backbones. The MViTv2 result is included from~\Cref{tab:closed-set-standard} for direct comparison.}
\vspace{-8pt}
\label{tab:skeleton-standard-backbones}
\setlength{\tabcolsep}{2pt}
\renewcommand{\arraystretch}{1.05}
\begin{adjustbox}{max width=\columnwidth}
\begin{tabular}{l l c c c c c c}
\toprule
\multirow{2}{*}{\textbf{Model}} & \multirow{2}{*}{\textbf{Input}}
& \multicolumn{3}{c}{\textbf{Classification}} & \multicolumn{3}{c}{\textbf{Retrieval}} \\
\cmidrule(lr){3-5} \cmidrule(lr){6-8}
& & \textbf{Top-1 (\%)} & \textbf{Top-3 (\%)} & \textbf{Top-5 (\%)} 
& \textbf{mAP (\%)} & \textbf{R-1 (\%)} & \textbf{R-5 (\%)} \\
\midrule
MViTv2~\cite{li2022mvitv2}
& Skel. 
& $97.84$ & $99.16$ & $99.30$
& $97.07$ & $97.64$ & $97.79$ \\
VideoMAEV2~\cite{wang2023videomae}
& Skel. 
& $25.07$ & $44.13$ & $53.49$
& $11.96$ & $24.78$ & $44.99$ \\
UniFormerV2~\cite{li2023uniformerv2}
& Skel. 
& $87.92$ & $95.53$ & $97.00$
& $55.38$ & $80.53$ & $93.22$ \\
\bottomrule
\end{tabular}
\end{adjustbox}
\end{table}

\subsection{Backbone-dependent Skeleton Compatibility}
% \noindent{\textbf{Backbone-dependent Skeleton Compatibility.}}
\Cref{tab:skeleton-standard-backbones} indicates that skeleton input is not equally effective across video backbones. MViTv2 remains highly robust under skeleton input, consistent with the observation in~\cref{ss_main_result}. UniFormerV2 also retains strong classification accuracy, although its retrieval performance is substantially weaker than that of MViTv2. In contrast, VideoMAEv2 degrades sharply across both classification and retrieval metrics. These results suggest that skeleton compatibility depends not only on the input regime itself, but also on how each backbone aggregates identity-relevant evidence from sparse spatio-temporal signals.

\subsection{Interpretation from Saliency Patterns}
\label{supp:explain-saliency}

A plausible interpretation is that different backbones aggregate identity evidence over the body at different spatial scales. This matters for skeleton input because skeletons preserve only sparse keypoints and their temporal progression, while removing the dense body envelope available in silhouettes. As a result, backbones that rely on broad, distributed body-level evidence may remain more compatible with skeletons, whereas backbones that depend more on localized regional evidence may be more sensitive to this sparsification.

As further illustrated in the backbone-level saliency comparison in~\cref{supp:backbone-saliency}, especially the left half of~\cref{fig:backbone-saliency}, MViTv2 exhibits a more spatially distributed response over the player body.~Its saliency covers lower-body, torso, and upper-body regions as the free-throw progresses.~UniFormerV2 shows a more region-focused pattern, with stronger emphasis on selected body parts and phase-specific configurations.~VideoMAEV2 appears the most localized, concentrating saliency on smaller regions rather than broadly aggregating evidence across the full body. The skeleton results in~\Cref{tab:skeleton-standard-backbones} broadly follow this ordering: MViTv2, which shows the most globally distributed saliency pattern, is also the most robust to skeleton input, whereas VideoMAEV2, which shows the most localized pattern, suffers the largest degradation.
\begin{figure}[t]
\renewcommand{\thefigure}{D1}
    \centering
    \begin{adjustbox}{max width=\columnwidth}
    \includegraphics{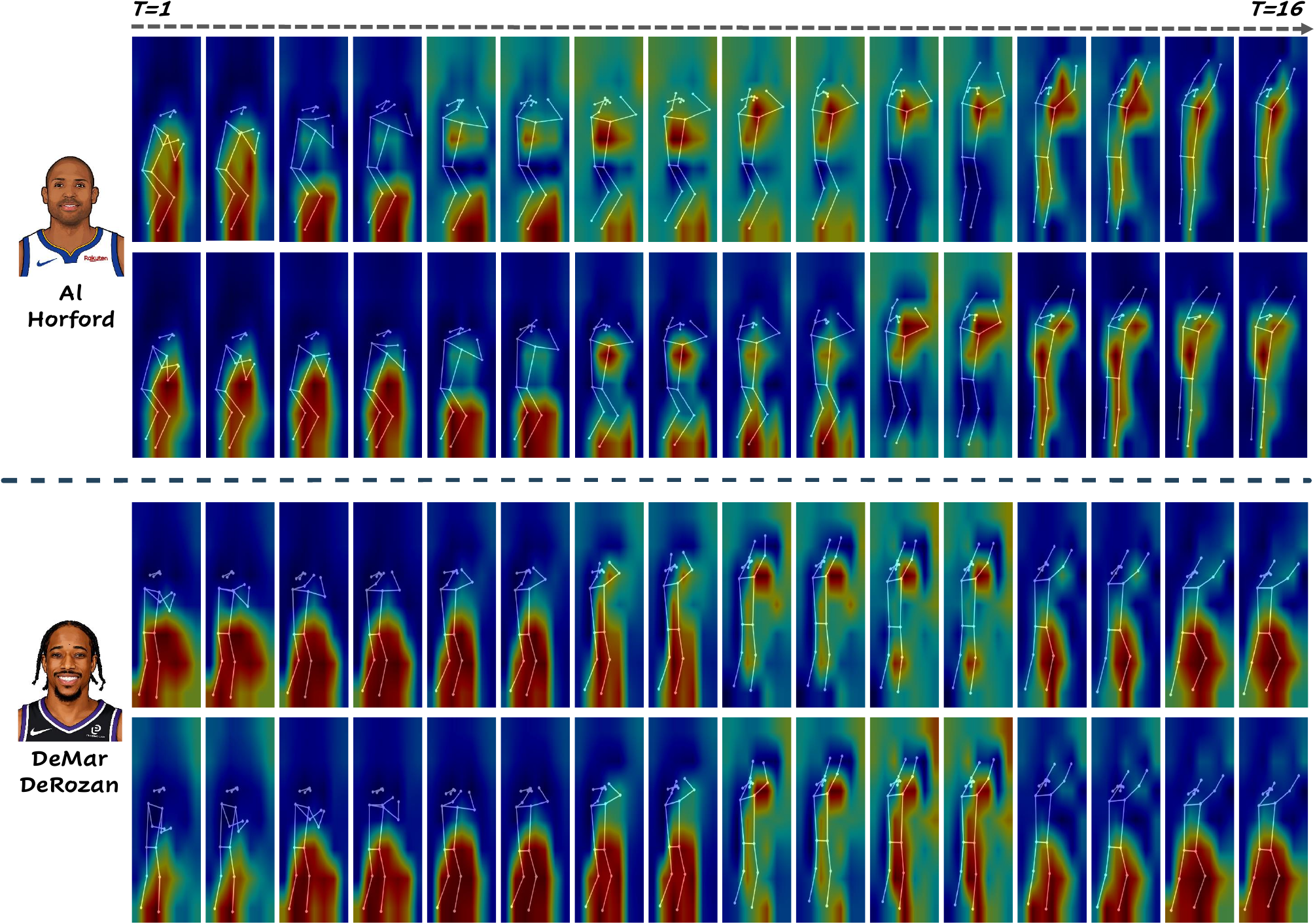}
    \end{adjustbox}
    \caption{Skeleton saliency maps across players and clips. For the same player, saliency patterns remain broadly consistent across clips, indicating within-identity stability.~Across players, the highlighted regions and temporal progression differ, suggesting identity-specific execution evidence.~The maps also show a relatively global response over the skeleton sequence, consistent with the stronger MViTv2 skeleton performance reported in~\cref{supp:skeleton-standard}.
    }
    \label{fig:skeleton-saliency}
    \vspace{-8pt}
\end{figure}

This difference is consistent with the input-regime design in~\cref{sec:dataset}. Silhouettes suppress face, jersey, and texture, but still preserve a dense body outline and rich spatial configuration. Skeletons further abstract the signal into sparse joints, which can weaken some execution-dependent cues discussed in~\cref{sec:protocol}, including phase-specific body configuration, relative limb placement, torso alignment, and localized coordination patterns. Under this view, skeletonization does not remove all identity evidence, but it changes which forms of evidence remain accessible to each architecture.

This interpretation also supports the observation in~\cref{ss_main_result} that silhouettes are generally more compatible with standard video backbones than skeletons. Silhouettes retain a denser representation of how the body is configured and moves through the routine, whereas skeletons provide a more compact and sparse abstraction. We therefore treat the saliency correspondence as a hypothesis consistent with the observed trends, not as independent causal evidence.

\begin{figure}[t]
\renewcommand{\thefigure}{E1}
    \centering
    \includegraphics[width=\linewidth]{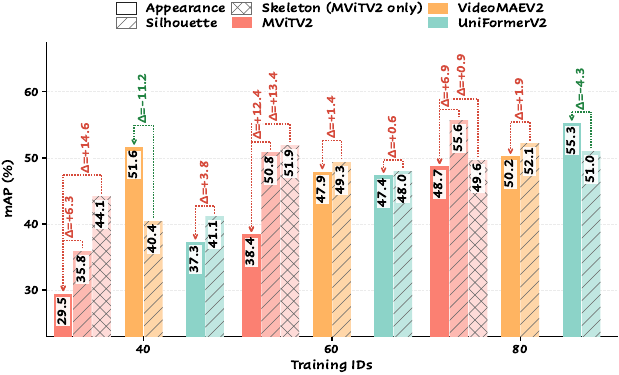}
    \caption{Open-set performance under the standard split. Appearance-suppressed regimes remain mostly ahead across training scales, showing that the trend observed under the appearance-disjoint setting also holds under a relatively easier protocol.}
    \label{fig:openset-standard}
    \vspace{-13pt}
\end{figure}
\section{Skeleton Saliency Analysis}
\label{supp:skeleton-attribution}

To further inspect the skeleton-input regime, we visualize CAM-based saliency maps for the MViTv2 skeleton model. This analysis complements the quantitative results in~\cref{supp:skeleton-standard} and follows the diagnostic criteria in~\cref{supp:inter-criteria}: the attributed evidence should be stable within an identity, distinct across identities, and aligned with the free-throw execution. 

As shown in~\cref{fig:skeleton-saliency}, the skeleton model produces consistent saliency patterns for the same player across different clips. For Al Horford, the highlighted evidence follows a similar temporal progression across examples, with attention broadly distributed over the lower body and torso in early frames and shifting toward the upper body near release. For DeMar DeRozan, the model follows a different progression, emphasizing a distinct pattern of lower-body, torso, and upper-body evidence across the sequence. These differences indicate that the model is not simply responding to a generic skeleton template shared by all players; rather, it highlights player-specific execution patterns within the same structured action.

This visualization is also consistent with the backbone-level observation in~\cref{supp:explain-saliency}. In particular, the saliency maps show a relatively global response over the skeleton sequence, rather than a narrowly localized focus on a single joint or region. Such distributed evidence may help explain why MViTv2 remains compatible with skeleton input: even after the dense body envelope is removed, the model can still aggregate identity-relevant cues across multiple body regions and phases. We therefore treat these maps as qualitative support for skeleton-based execution evidence, while maintaining the diagnostic scope discussed in~\cref{supp:inter-criteria}.
\begin{figure}[t]
\renewcommand{\thefigure}{F1}
    \centering
    \includegraphics[width=\columnwidth]{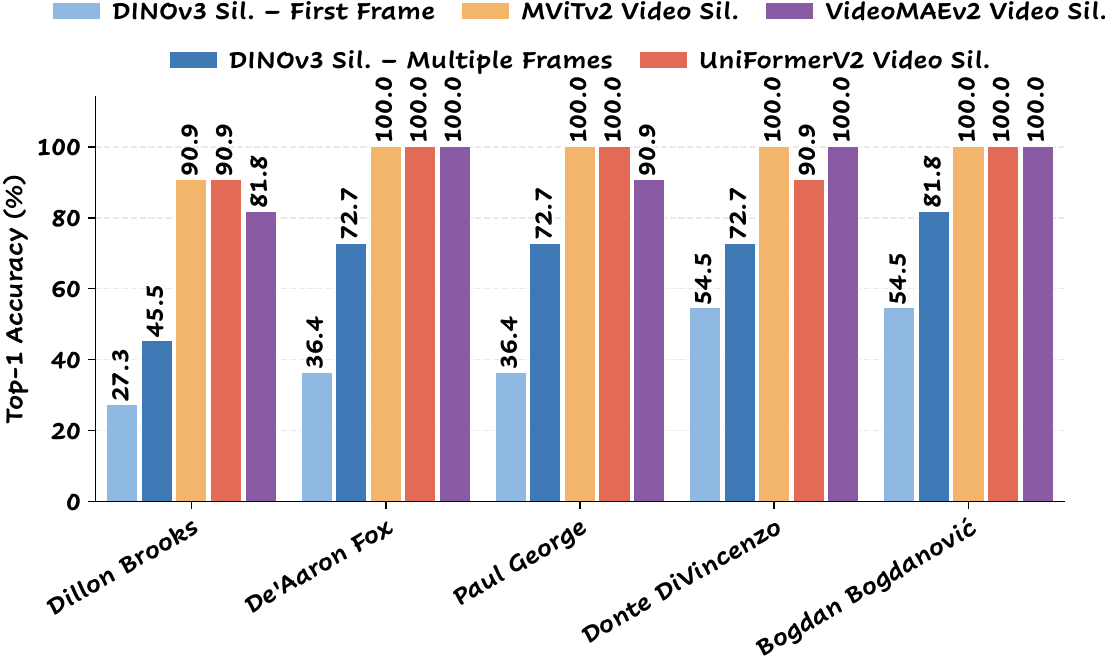}
    \caption{Per-identity static-frame analysis for silhouette inputs. Multiple frames improve over a single frame, while video models further exploit temporal execution cues. }
    \label{fig:static-frame-silhouette}
    \vspace{-8pt}
\end{figure}
\begin{figure}[t]
\renewcommand{\thefigure}{F2}
    \centering
    \includegraphics[width=\columnwidth]{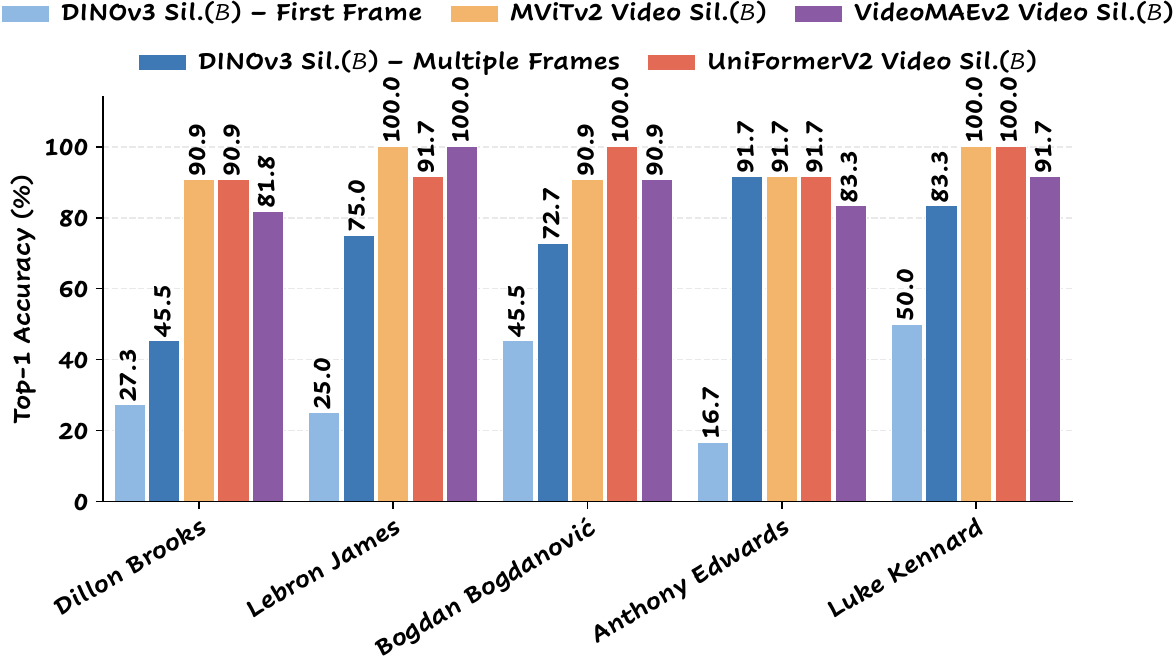}
    \caption{ Per-identity static-frame analysis for contour-degraded silhouettes. Video models overally remain strong after contour degradation, supporting the role of temporal execution cues beyond static boundary information. }
    \label{fig:static-frame-silhouette-blur}
    \vspace{-15pt}
\end{figure}
\begin{figure}[t]
\renewcommand{\thefigure}{F3}
    \centering
    \includegraphics[width=\columnwidth]{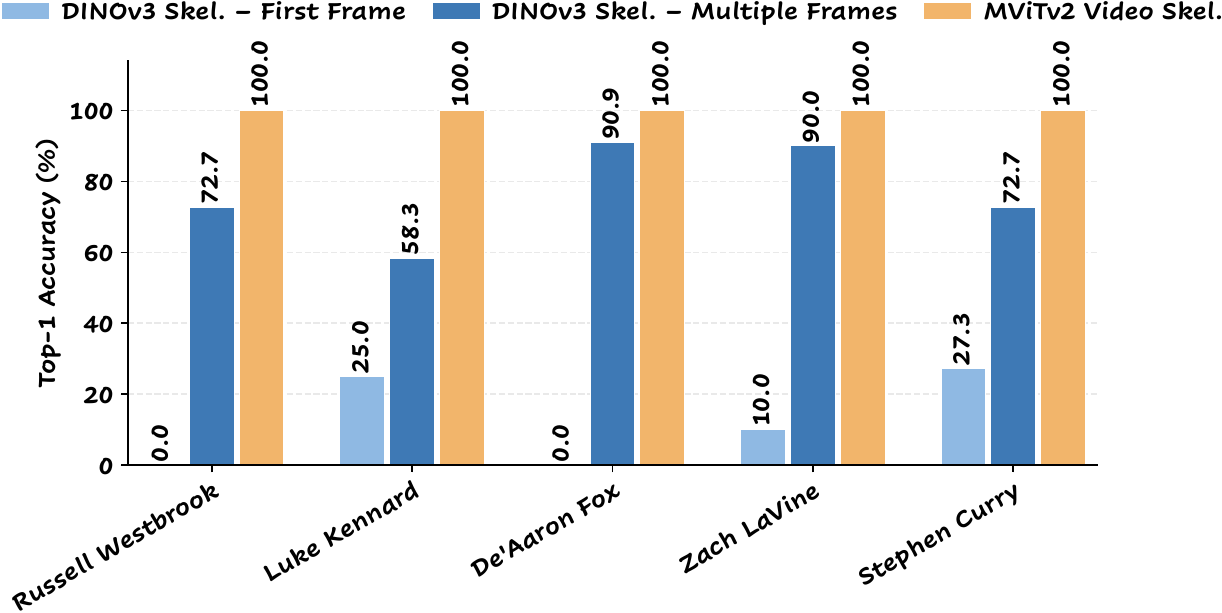}
    % \vspace{-10pt}
    \caption{Per-identity static-frame analysis for skeleton inputs. Video modeling provides stronger identity evidence than isolated or independently processed static skeleton frames. }
    \label{fig:static-frame-skeleton}
    \vspace{-15pt}
\end{figure}
\section{Standard Open-set Results}
\label{supp:standard-open-set}
In~\cref{ss_main_result}, we evaluate open-set recognition under the appearance-disjoint split, where training and test clips differ in jersey appearance.~This protocol is intentionally strict: it reduces the usefulness of static appearance and tests whether appearance-suppressed representations transfer to unseen identities.~Here, we add a complementary open-set evaluation under the standard split, where the held-out identities are still unseen but the train/test appearance distributions are not explicitly separated by jersey appearance.

As shown in~\cref{fig:openset-standard}, the overall trend remains consistent with the appearance-disjoint open-set results in~\cref{ss_main_result}.~Although the standard split is less challenging and therefore yields smaller gains, appearance-suppressed regimes still outperform appearance in most settings. Silhouettes remain consistently competitive across backbones, and the MViTv2 skeleton results also stay strong when available.~This supports the observation in~\cref{ss_main_result} that identity-relevant execution cues are not limited to the appearance-disjoint protocol. These results should be read as complementary evidence rather than a replacement for the appearance-disjoint evaluation. The appearance-disjoint split remains the stricter test of robustness to appearance shifts, while the standard split shows that the same preference for appearance-suppressed representations largely persists even when the evaluation protocol is relatively easier.

\section{Static Evidence and Temporal Ordering}
\label{supp:static-temporal}

\cref{sec:further-exploration} asks whether identity-specific motion signatures can be recovered from static frames, or whether ordered video provides additional evidence. We complement that analysis with two controlled probes. The first compares frame-based image representations with ordered video representations under appearance-suppressed regimes, testing how much identity evidence is already present in isolated or independently processed phases. The second disrupts temporal order by shuffling frames within each video, testing whether video models rely on the ordered progression of the free-throw routine rather than only on the unordered set of body configurations. Together, these analyses clarify the role of temporal structure: static poses provide partial identity evidence, but ordered execution across phases remains a stronger cue for motion-based identity recognition.

\subsection{Frame-based vs. Video-based Recognition}
\label{supp:frame-video}

The results in~\cref{sec:further-exploration} show that static frames contain partial identity evidence, but ordered video provides stronger recognition. Here, we further inspect this gap at the identity level. Rather than showing randomly selected players, we focus on identities for which video modeling provides the clearest benefit over frame-based recognition. This allows us to visualize cases where isolated or independently processed poses are insufficient, but ordered execution across the free-throw sequence remains discriminative.

Displayed identities are selected by comparing the average Top-1 accuracy of image-based DINOv3 representations across the first-frame and multiple-frame settings with the average Top-1 accuracy of the available video models. For each appearance-suppressed regime, we show the five identities with the largest improvement from frame-based to video-based recognition. For skeleton input, only MViTv2 is included as the available video model.

As shown in~\cref{fig:static-frame-skeleton,fig:static-frame-silhouette,fig:static-frame-silhouette-blur}, static frames can already support identity recognition in some cases, especially when multiple phases are observed. This is consistent with the discussion in~\cref{sec:protocol}: some execution-dependent cues appear within individual phases, such as set posture, release pose, or follow-through configuration. However, video-based models remain stronger for the selected identities, indicating that these cues are not fully captured by isolated static configurations. Instead, identity evidence is distributed across the free-throw sequence, and ordered temporal relationships between phases provide additional discriminative information.

\subsection{Temporal Shuffling}
% Put these color definitions in the preamble if not already defined.
\definecolor{appblue}{RGB}{235,245,255}
\definecolor{silred}{RGB}{255,240,240}

\begin{table}[t]
\renewcommand{\thetable}{F1}
\centering
\caption{Impact of temporal shuffling ($\mathcal{S}$) on VideoMAEV2 and UniFormerV2. Shuffling preserves the frames but removes their temporal order; the larger degradation for silhouette inputs indicates reliance on ordered execution dynamics.}
\label{tab:temporal_shuffling}
\begin{adjustbox}{max width=\columnwidth}
\begin{tabular}{llllllll}
\toprule
\multirow{2}{*}{\textbf{Model}} &
\multirow{2}{*}{\textbf{Input}} &
\multicolumn{3}{c}{\textbf{Classification}} &
\multicolumn{3}{c}{\textbf{Retrieval}} \\
\cmidrule(lr){3-5} \cmidrule(lr){6-8}
& & \textbf{Top-1 (\%)} & \textbf{Top-3 (\%)} & \textbf{Top-5 (\%)}
& \textbf{mAP (\%)} & \textbf{R-1 (\%)} & \textbf{R-5 (\%)} \\
\midrule
\multirow{4}{*}{VideoMAEV2~\cite{wang2023videomae}}
& App. & $98.04$ & $98.95$ & $99.16$ & $97.93$ & $98.53$ & $98.53$ \\
& \cellcolor{appblue}App.$+\mathcal{S}$
& \cellcolor{appblue}$96.93$\,\diffdown{1.11}
& \cellcolor{appblue}$98.74$\,\diffdown{0.21}
& \cellcolor{appblue}$99.09$\,\diffdown{0.07}
& \cellcolor{appblue}$96.10$\,\diffdown{1.83}
& \cellcolor{appblue}$97.33$\,\diffdown{1.20}
& \cellcolor{appblue}$97.88$\,\diffdown{0.65} \\
& Sil. & $95.46$ & $98.67$ & $99.09$ & $95.58$ & $97.64$ & $98.08$ \\
& \cellcolor{silred}Sil.$+\mathcal{S}$
& \cellcolor{silred}$43.78$\,\diffdown{51.68}
& \cellcolor{silred}$60.68$\,\diffdown{37.99}
& \cellcolor{silred}$69.13$\,\diffdown{29.96}
& \cellcolor{silred}$31.83$\,\diffdown{63.75}
& \cellcolor{silred}$55.27$\,\diffdown{42.37}
& \cellcolor{silred}$76.19$\,\diffdown{21.89} \\
\midrule
\multirow{4}{*}{UniFormerV2~\cite{li2023uniformerv2}}
& App. & $97.49$ & $99.02$ & $99.30$ & $89.35$ & $97.20$ & $98.97$ \\
& \cellcolor{appblue}App.$+\mathcal{S}$
& \cellcolor{appblue}$93.85$\,\diffdown{3.64}
& \cellcolor{appblue}$97.63$\,\diffdown{1.39}
& \cellcolor{appblue}$98.32$\,\diffdown{0.98}
& \cellcolor{appblue}$73.60$\,\diffdown{15.75}
& \cellcolor{appblue}$87.85$\,\diffdown{9.35}
& \cellcolor{appblue}$96.12$\,\diffdown{2.85} \\
& Sil. & $95.67$ & $98.25$ & $98.95$ & $85.95$ & $95.13$ & $97.49$ \\
& \cellcolor{silred}Sil.$+\mathcal{S}$
& \cellcolor{silred}$80.80$\,\diffdown{14.87}
& \cellcolor{silred}$91.06$\,\diffdown{7.19}
& \cellcolor{silred}$94.20$\,\diffdown{4.75}
& \cellcolor{silred}$53.14$\,\diffdown{32.81}
& \cellcolor{silred}$77.42$\,\diffdown{17.71}
& \cellcolor{silred}$89.88$\,\diffdown{7.61} \\
\bottomrule
\end{tabular}
\end{adjustbox}
\end{table}
\label{supp:temporal-shuffling}

The frame-based analysis tests whether static phase information is sufficient. We next test whether temporal order itself matters by training and evaluating VideoMAEV2 and UniFormerV2 under the standard closed-set split after randomly permuting all frames within each clip. This preserves the same visual observations while disrupting the ordered progression of the free-throw routine.

As shown in~\Cref{tab:temporal_shuffling}, temporal shuffling affects appearance and silhouette models differently. Appearance models degrade only moderately, suggesting that they can still rely on static visual evidence. In contrast, silhouette models suffer substantially larger drops, especially for VideoMAEV2. Since silhouette inputs suppress face, jersey, and texture, this degradation indicates that the models are not merely using an unordered collection of body configurations; they also rely on the temporal ordering of execution cues. This complements the frame-based analysis and further supports the conclusion in~\cref{sec:further-exploration}: static poses contain partial identity evidence, but ordered motion provides a more robust basis for probing identity-specific motion signatures.

\section{Representation Evidence Across Regimes}
\label{supp:representation-evidence}

The analyses in~\cref{ss_main_result,sec:main-analysis,sec:further-exploration} show that appearance and appearance-suppressed models achieve recognition through different evidence: appearance models remain vulnerable to appearance shifts, whereas silhouette models preserve stronger robustness and attend to phase-aligned execution cues. Here, we provide a complementary representation-level view using UniFormerV2 as a representative backbone. Instead of asking where the model attends, we ask how the learned feature space organizes clips under different input regimes. We use nearest-neighbor retrieval and embedding visualization to inspect whether appearance and silhouette inputs induce different identity neighborhoods.

\subsection{Nearest-neighbor Retrieval}
\label{supp:retrieval-neighborhoods}
\begin{figure}[t]
\renewcommand{\thefigure}{G1}
    \centering
    \includegraphics[width=0.75\columnwidth]{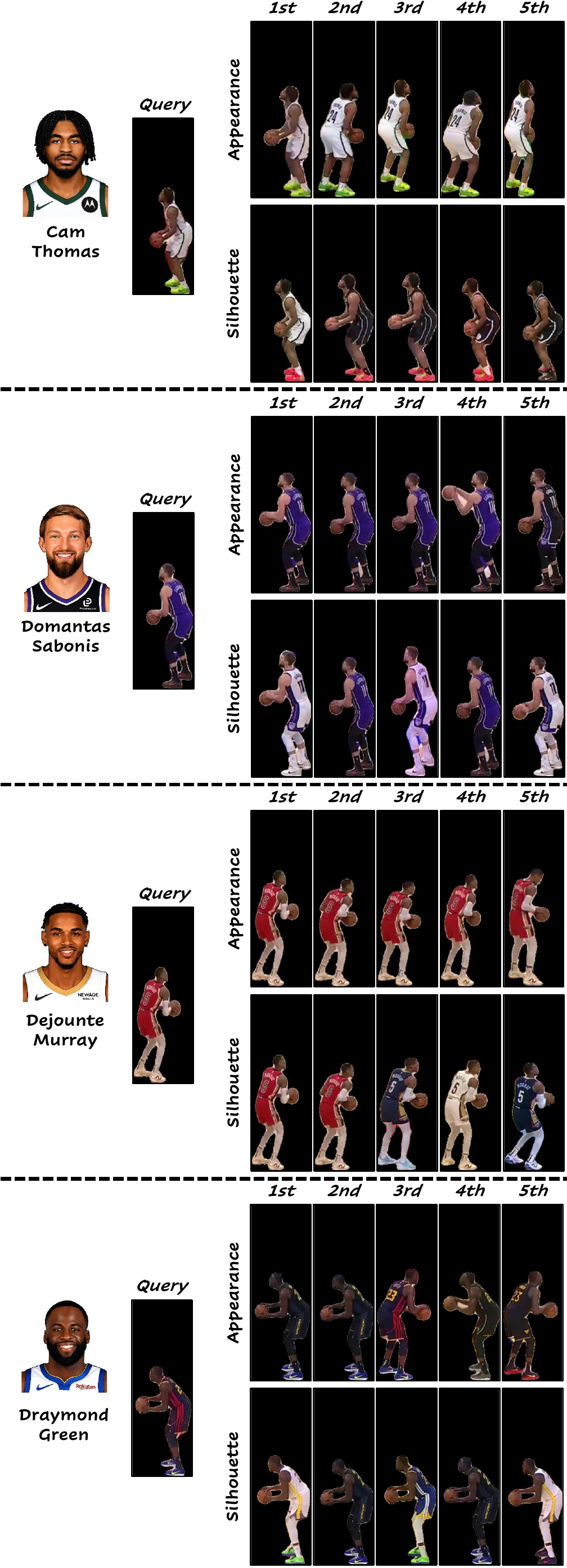}
    \caption{
    Top-5 nearest gallery clips retrieved under appearance and silhouette inputs.
    Appearance retrieval tends to preserve visual similarity, while silhouette retrieval returns appearance-diverse clips.
    }
    \label{fig:retrieval-neighbors}
    \vspace{-15pt}
\end{figure}

We first compare retrieval neighborhoods under full-appearance and silhouette input. For each query clip, we retrieve the Top-5 nearest gallery clips from the learned embedding space. For silhouette retrieval, we display the corresponding full-appearance videos of the retrieved clips only for visual inspection; retrieval itself is performed using silhouette features.

As shown in~\cref{fig:retrieval-neighbors}, appearance-based retrieval often returns gallery clips with similar visual appearance, such as jersey color or texture, even when these cues are not the intended focus of our diagnostic protocol. In contrast, silhouette-based retrieval produces neighbors with more diverse visual appearances, since face, jersey, and texture are removed from the input. This difference is consistent with the observation in~\cref{ss_main_result,sec:main-analysis}: full-appearance models can exploit static shortcuts, while appearance-suppressed models rely more on how the body is configured and moves through the free-throw routine.

\subsection{Embedding-space Visualization}
\label{supp:embedding-visualization}
\begin{figure}[t]
\renewcommand{\thefigure}{G2}
    \centering
    \includegraphics[width=\columnwidth]{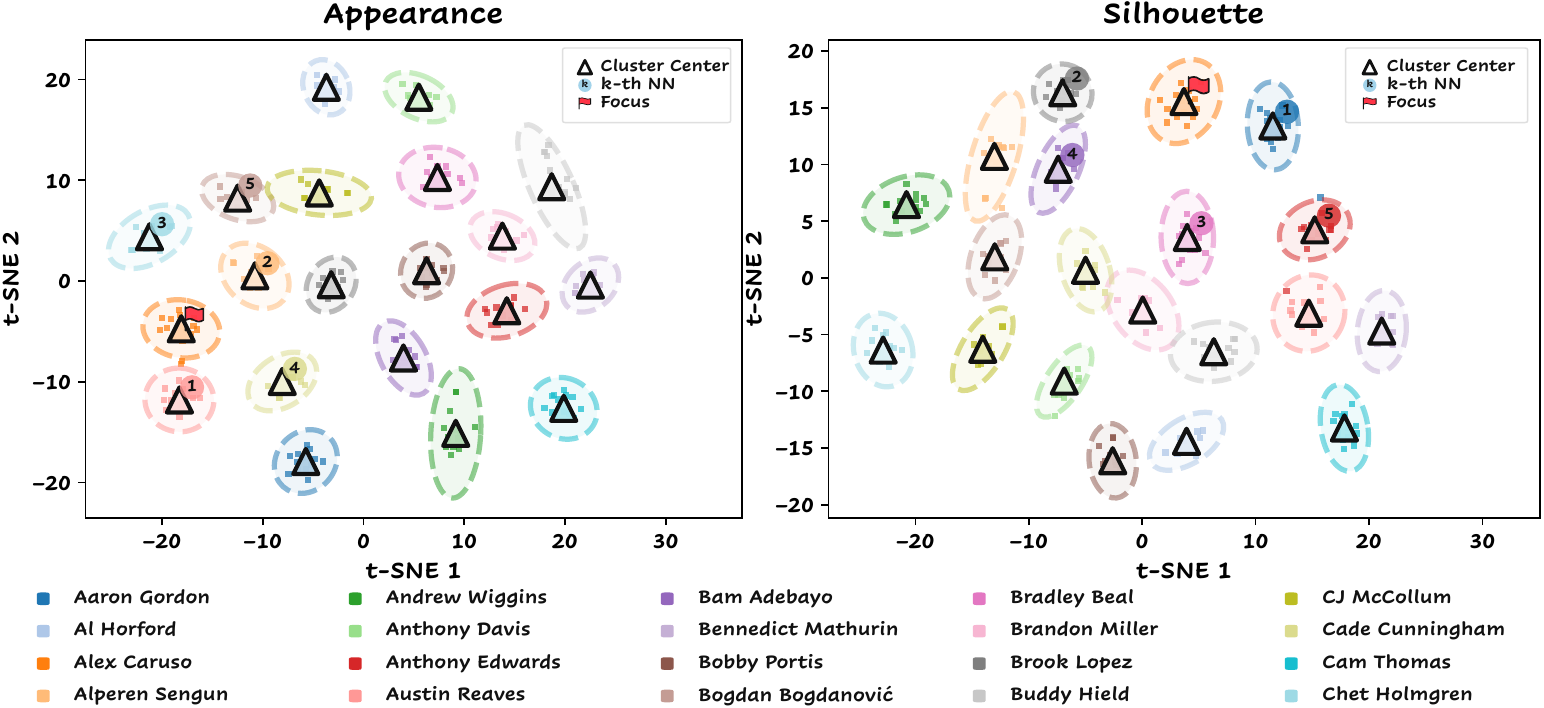}
    \caption{
    t-SNE visualization of identity-level cluster centers under appearance and silhouette inputs.
    The red flag marks the same focus identity, and numbers indicate its Top-5 nearest neighboring identity centers in each feature space.
    The changed neighbor ordering shows that appearance and silhouette models organize identity evidence differently.
    }
    \label{fig:tsne-representation}
\end{figure}

We further visualize the learned embedding spaces using t-SNE. Each marker denotes the cluster center of one identity, and colors indicate player identity. To compare neighborhood structure across regimes, we mark one focus identity and annotate its Top-5 nearest neighboring identity centers in the appearance and silhouette feature spaces.

As shown in~\cref{fig:tsne-representation}, the nearest-neighbor structure changes substantially after appearance is suppressed. In the appearance feature space, the focus identity is surrounded by one set of neighboring identities, whereas in the silhouette feature space its closest neighbors are reorganized and reordered. This indicates that the two regimes do not simply produce the same identity geometry under different input channels. Instead, appearance and silhouette inputs induce different neighborhood structures, suggesting that the learned representations emphasize different cue families.

This visualization complements the retrieval examples in~\cref{supp:retrieval-neighborhoods}.~Appearance-based representations tend to organize clips according to static visual similarity, while silhouette-based representations are more consistent with appearance-suppressed execution evidence. We therefore treat the t-SNE discrepancy as representation-level support for the cue shift observed in~\cref{ss_main_result,sec:main-analysis}, rather than as standalone proof of motion-based recognition.
\begin{figure*}[t]
\renewcommand{\thefigure}{I1}
    \centering
    \includegraphics[width=\textwidth]{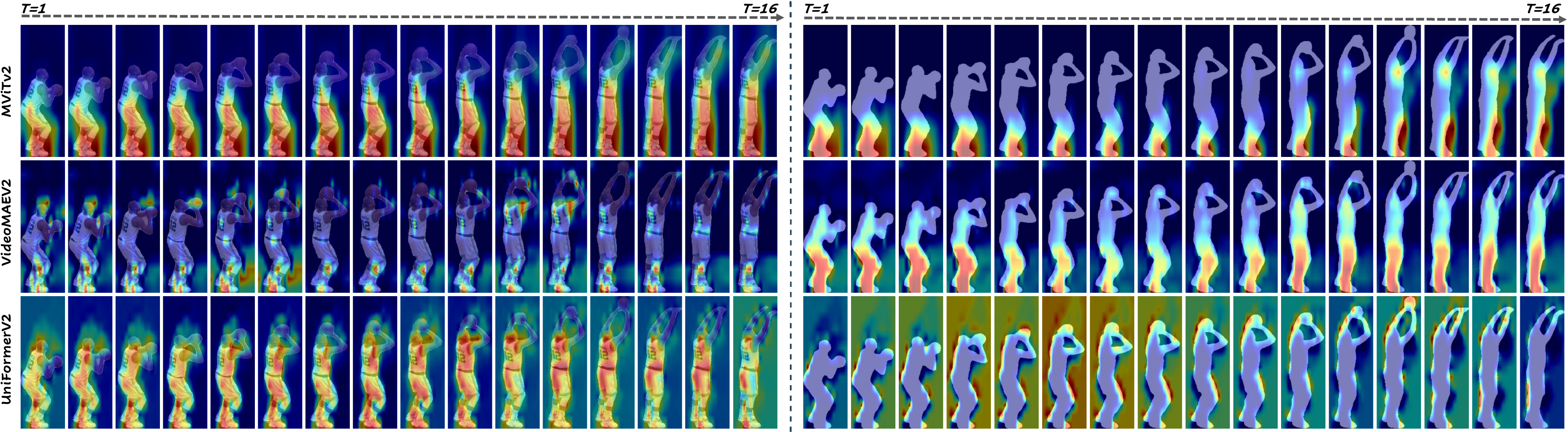}
    \caption{
    Backbone-dependent saliency patterns under appearance and silhouette inputs. For the same identity and sequence, MViTv2 shows a more spatially distributed response, whereas VideoMAEV2 and UniFormerV2 exhibit more localized or region-focused patterns. These qualitative differences support the view that backbones aggregate execution-dependent identity evidence at different spatial scales.
    }
    \label{fig:backbone-saliency}
\end{figure*}
\section{Stable and Distinctive Execution Evidence}
\begin{figure*}[p]
\renewcommand{\thefigure}{H1}
    \centering
    \includegraphics[width=\linewidth, height=0.9\textheight, keepaspectratio]{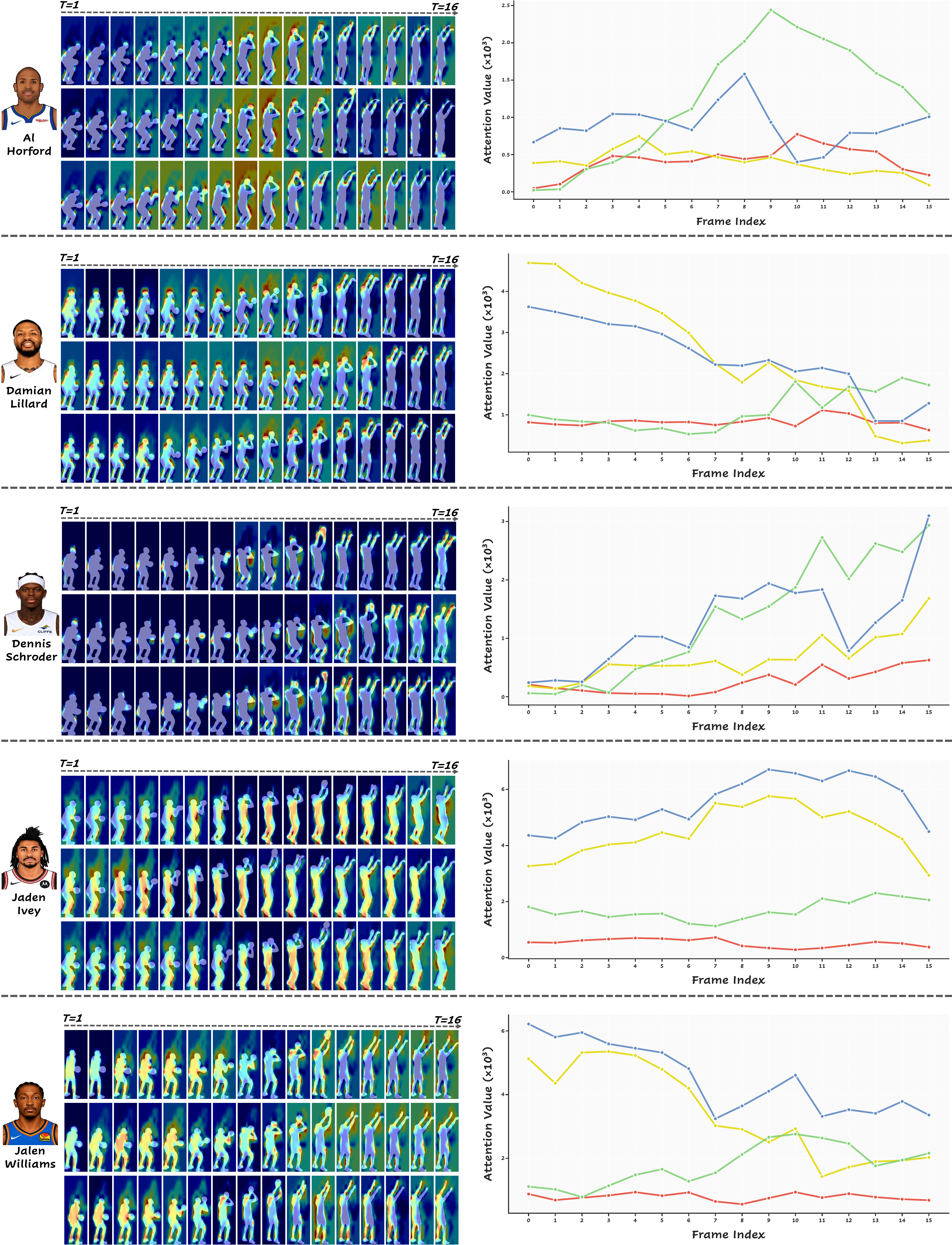}
    \caption{
    Stable and distinctive execution evidence across five identities.
    For each player, the left panel shows CAM-based saliency maps over multiple clips through the sequence, while the right panel reports pose-guided quantitative saliency curves over anatomical regions: \textcolor{red}{Head}, \textcolor[HTML]{E2D909}{Torso}, \textcolor[HTML]{82D177}{Arms}, and \textcolor[HTML]{5E89C2}{Legs}.
    The saliency patterns remain consistent within each identity but differ across identities, supporting the view that motion signatures emerge from compositional, phase-aligned execution cues rather than a single localized body part.
    }
    \label{fig:signature-stability}
\end{figure*}

\label{supp:signature-stability}

The saliency analysis in~\cref{sec:main-analysis} suggests that appearance-suppressed models attend to execution-dependent evidence rather than static visual shortcuts. Here, we provide extended qualitative and quantitative support for the diagnostic criteria introduced in~\cref{supp:saliency}: the evidence should be stable within an identity, distinct across identities, and aligned with interpretable phases of the free-throw routine. We visualize five representative players: \textbf{\textit{Al Horford}}, \textbf{\textit{Damian Lillard}}, \textbf{\textit{Dennis Schroder}}, \textbf{\textit{Jaden Ivey}}, and \textbf{\textit{Jalen Williams}}. As shown in~\cref{fig:signature-stability}, each row contains CAM-based saliency maps across multiple clips and pose-guided temporal saliency curves for the same identity.

\subsection{Qualitative Stability and Distinctiveness}
\label{supp:qualitative-stability}

Across clips of the same player, the silhouette model repeatedly highlights similar regions at corresponding phases, indicating that the saliency patterns are not dominated by incidental frame-level noise. Across players, however, the highlighted regions and their temporal progression differ. This supports the view that the model does not rely on a single universal cue shared by all free-throws, but instead combines multiple execution-dependent cues into identity-specific motion signatures.

For \textbf{\textit{Al Horford}}, the model consistently emphasizes lower-body and torso evidence during the load and rise phases, followed by increased attention to the shooting arm and upper-body finish near release. For \textbf{\textit{Damian Lillard}}, saliency appears earlier and more strongly around the compact preparatory posture, including the lower body, torso, and set-point configuration, before shifting toward the shooting arm during release. For \textbf{\textit{Dennis Schroder}}, the highlighted evidence is initially weaker and more localized, then becomes more prominent around the torso, hip, and arm trajectory as the motion progresses. For \textbf{\textit{Jaden Ivey}}, the maps show a broader response over the body, consistent with a more global coordination pattern rather than a single localized joint cue. For \textbf{\textit{Jalen Williams}}, the model also uses broad body evidence, but with a different temporal profile, including sustained attention to standing stability and upper-limb behavior during the later shooting phases.

These player-wise patterns indicate that identity-specific motion signatures are compositional. They are not defined by one isolated body part, such as only the arm or only the feet. Instead, the model appears to combine phase-specific posture, lower-body loading, torso alignment, limb coordination, and follow-through configuration. This interpretation is consistent with our diagnostic design in~\cref{sec:protocol}, where execution-dependent evidence can appear within individual phases and also unfold across transitions between phases.
\subsection{Pose-guided Quantitative Saliency Curves}
\label{supp:pose-guided-curves}

To quantitatively complement the frame-level saliency maps, we aggregate saliency values over pose-guided anatomical regions. Using estimated keypoints, we group the body into head, torso, arms, and legs, and compute region-level saliency over the sequence. The resulting curves in~\cref{fig:signature-stability} provide a temporally resolved quantitative summary of how the model's focus evolves across the free-throw routine.

These quantitative profiles reinforce the qualitative observations. First, they show phase-aligned temporal structure: saliency shifts across body regions as the action moves from set posture to rise, release, and follow-through. Second, they show identity-specific profiles: different players exhibit different saliency trajectories even though they perform the same action. Third, they are consistent with the spatial maps, indicating that the pose-guided curves summarize the same evidence observed in the CAM visualizations rather than introducing an unrelated measurement.

Taken together, the qualitative saliency maps and quantitative pose-guided curves support the central claim that appearance-suppressed models recover stable, distinctive, and interpretable execution evidence. The resulting motion signature is best understood as a structured combination of body-region cues and phase-specific dynamics, rather than as a single localized visual feature. As in~\cref{supp:saliency}, we interpret these results as diagnostic evidence for model behavior, not as standalone proof.

\section{Backbone-dependent Saliency Patterns}
\label{supp:backbone-saliency}
We finally compare saliency patterns across the three video backbones under the same identity and sequence. This analysis complements the backbone-sensitivity discussion in~\cref{supp:skeleton-standard}: if different architectures aggregate identity evidence at different spatial scales, their saliency maps should reveal different patterns even when the input sequence and target identity are fixed.

As shown in~\cref{fig:backbone-saliency}, the three backbones exhibit visibly different saliency distributions. MViTv2 produces a relatively broad and spatially distributed response, with highlighted regions spanning the lower body, torso, and upper body as the free-throw progresses. VideoMAEV2 shows a more localized pattern, especially under appearance input, where attention concentrates on smaller regions rather than covering the full body trajectory. UniFormerV2 exhibits an intermediate but more region-focused behavior, with saliency often concentrated around selected body parts or phase-specific configurations. These differences are also visible under silhouette input, where static appearance cues are removed. MViTv2 continues to aggregate evidence over a broader body region across time, whereas VideoMAEV2 and UniFormerV2 show more concentrated responses around local execution cues. This observation is consistent with the hypothesis in~\cref{supp:explain-saliency}: backbones that aggregate evidence more globally may be more robust when the input is reduced to sparse skeletons, while models that rely more heavily on localized regional evidence may be more sensitive to the loss of dense spatial structure.

Importantly, these patterns do not imply that one backbone uses a single cue and another uses a different single cue. Rather, they suggest that identity-specific motion signatures can be represented through different architectural strategies. The evidence remains execution-dependent, but the spatial scale at which each backbone organizes that evidence differs. We therefore treat this comparison as qualitative support for the broader claim that motion signatures are compositional and architecture-dependent, rather than tied to one fixed saliency pattern.

\end{document}